\documentclass[10pt,twocolumn,letterpaper]{article}

\usepackage[pagenumbers]{cvpr} 
\usepackage[accsupp]{axessibility}
\usepackage{commands}
\usepackage{color, colortbl}
\usepackage{xcolor}
\usepackage{pifont}
\usepackage{subfloat}
\usepackage{booktabs}
\usepackage{diagbox}
\usepackage{multirow}
\usepackage{makecell}
\usepackage{amsmath}
\usepackage{algorithm}
\usepackage{algpseudocode}
\usepackage{cuted}
\usepackage{pifont}
\newcommand{\cmark}{\ding{51}}%
\definecolor{Gray}{gray}{0.9}
\DeclareUnicodeCharacter{03CC}{\'o}
\definecolor{cvprblue}{rgb}{0.21,0.49,0.74}
\usepackage[pagebackref,breaklinks,colorlinks,allcolors=cvprblue]{hyperref}

\def\paperID{13848} 
\def\confName{CVPR}
\def\confYear{2025}

\title{Learning Phase Distortion with Selective State Space Models \\
for Video Turbulence Mitigation}

\author{Xingguang Zhang \ \ \  Nicholas Chimitt \ \ \ Xijun Wang \ \ \ Yu Yuan \ \ \ Stanley H. Chan\\
School of Electrical and Computer Engineering, Purdue University \\
{\tt\small \{zhan3275, nchimitt, wang6661, yuan418, stanchan\}@purdue.edu}
}

\begin{document}
\maketitle
\begin{abstract}
Atmospheric turbulence is a major source of image degradation in long-range imaging systems. Although numerous deep learning-based turbulence mitigation (TM) methods have been proposed, many are slow, memory-hungry, and do not generalize well. In the spatial domain, methods based on convolutional operators have a limited receptive field, so they cannot handle a large spatial dependency required by turbulence. In the temporal domain, methods relying on self-attention can, in theory, leverage the lucky effects of turbulence, but their quadratic complexity makes it difficult to scale to many frames. Traditional recurrent aggregation methods face parallelization challenges.

In this paper, we present a new TM method based on two concepts: (1) A turbulence mitigation network based on the Selective State Space Model (MambaTM). MambaTM provides a global receptive field in each layer across spatial and temporal dimensions while maintaining linear computational complexity. (2) Learned Latent Phase Distortion (LPD). LPD guides the state space model. Unlike classical Zernike-based representations of phase distortion, the new LPD map uniquely captures the actual effects of turbulence, significantly improving the model’s capability to estimate degradation by reducing the ill-posedness. Our proposed method exceeds current state-of-the-art networks on various synthetic and real-world TM benchmarks with significantly faster inference speed. The code is available at \href{https://xg416.github.io/MambaTM}{\textcolor{pink}{https://xg416.github.io/MambaTM}}.
\end{abstract}    
\section{Introduction}
\label{sec:intro}
Images captured from long-range distances often suffer from degradation caused by atmospheric turbulence. The spatiotemporal varying pixel displacement and blur introduced by the accumulation of wavefront phase distortion over long distance \cite{Fried_1965_a} create an unsatisfying visual effect and severely degrade downstream vision tasks such as detection or recognition that rely on the captured image \cite{cornett2023expanding, zhang2024source}. To solve this problem, deep learning-based turbulence mitigation (TM) methods have been developed recently with synthetic datasets produced by physics-based \cite{chimitt2022real, Mao_2021_a, Chimitt_2020_a, chan2023computational} or visual effect-based \cite{Jin_2021_a, saha2024turb, Anantrasirichai_2022_a} simulators.

Although recent data-driven video TM methods \cite{zhang2024spatio, Zhang_2022_a, Jin_2021_a} have shown impressive generalization capabilities, they heavily depend on training datasets and lack an interpretable understanding of turbulence degradation dynamics. In single-frame TM methods, to inject degradation awareness in training and improve adaptivity during training, \cite{Mao_2022_a, wang2024real, Jaiswal_2023_ICCV} propose to learn a re-degradation function that leverages both physics-based simulators and real-world images. However, the re-degradation function in \cite{Mao_2022_a, wang2024real} lacks clear physics-based interpretation. In contrast, \cite{Jaiswal_2023_ICCV} used a differentiable simulation engine \cite{chimitt2022real} to incorporate turbulence properties into the network. Despite this, adapting this approach to video TM networks is challenging. First, it requires knowledge of degradation parameters during training, limiting its applicability to datasets lacking such information. Second, the simulator \cite{chimitt2022real} uses large kernel depth-wise convolution, which is slow in an efficient restoration pipeline. Third, the blur kernel size in \cite{chimitt2022real} varies with turbulence conditions and is not differentiable. Finally, \cite{chimitt2022real} relies on Zernike polynomials \cite{noll1976zernike} to represent degradation, but estimating pixel-wise Zernike coefficients from a degraded image is highly ill-posed since different Zernike coefficient fields can produce the same degradation pattern.

To efficiently impose the interpretable degradation awareness, we proposed to reparameterize the physics-based turbulence simulator \cite{chimitt2022real} with a conditional variational autoencoder \cite{vae} (VAE). Specifically, our VAE encodes the phase distortion (PD), represented by the classical Zernike coefficient random field, into a latent map and conditionally decodes it into spatially varying blur patterns based on the input Zernike coefficients. This approach bypasses the undifferentiable kernel size and the slow large-kernel depth-wise convolution. Additionally, the space of all possible Zernike random fields corresponding to the same blur pattern is mapped to a more tractable Gaussian distribution. The mean and variance of this distribution, referred to as the latent phase distortion (LPD) map, uniquely determine the blur pattern and can be estimated by the restoration network.

With the learned LPD, we can jointly train degradation estimation and turbulence mitigation to improve the TM network's degradation awareness. However, video TM networks typically require a large spatiotemporal receptive field to capture the stochastic characteristics of degradation \cite{zhang2024spatio, cai2023convrt}. Joint training further increases the computational budget for both training and inference, presenting a challenge for efficient network design. Recently, Selective State Space Models (SSMs) \cite{gu2023mamba, dao2024mamba2} have shown advantages in various computer vision tasks \cite{visionmamba_2024, li2024videomamba}, including image and video restoration \cite{guo2024mambair, wu2024rainmamba}, due to their linear complexity and global receptive field over sequence length. Inspired by this, we apply the selective state space model (Mamba) to turbulence mitigation and propose MambaTM, which jointly estimates the LPD and restores videos affected by turbulence. Additionally, we use the learned latent phase distortion as a reference to guide state space construction in SSM, termed \emph{guided SSM}, to enhance our network's adaptivity. In summary, we offer the following contributions:
 

\begin{itemize}
    
    \item We propose a reparameterization trick to transform the Zernike-based representation of the turbulence degradation to a latent phase distortion (LPD) representation. Turbulence simulation with the LPD is $50 \times$ faster than the state-of-the-art turbulence simulator while preserving its physics property.

    \item Coupled with the LPD simulator, we present a variational framework to jointly estimate the turbulence degradation and mitigate the turbulence.

    \item We propose the first Mamba-based network, MambaTM, for the video turbulence mitigation problem. Specifically, we propose the phase-distortion guided Mamba block to facilitate degradation-aware turbulence mitigation and enhance the adaptivity of the network.
    
    \item Extensive experiments across multiple synthetic and real-world TM benchmarks demonstrate our method achieves state-of-the-art reconstruction quality while enjoying significantly faster inference speed than other approaches.
\end{itemize}
\section{Related Works}
\label{sec:related_works}
\begin{figure*}[h]
    \centering
    \includegraphics[width = 0.98\linewidth]{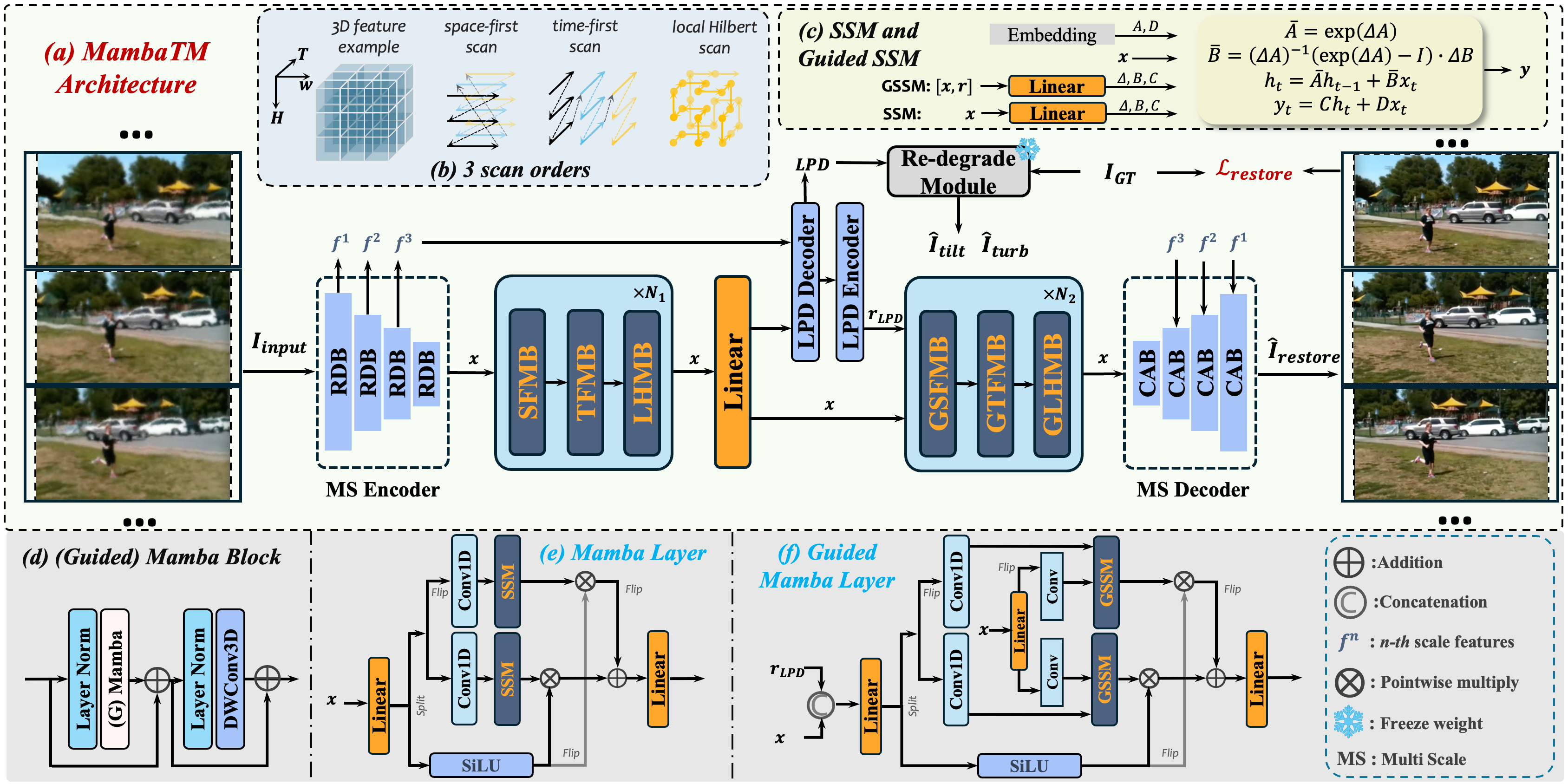}
    \caption{The proposed MambaTM network. The RDB means residual dense block \cite{Zhang_2021_a}, and CAB denotes the channel attention blocks \cite{Zhang_2018_a}. SFMB, TFMB, LHMB means space-first, time-first, and local Hilbert Mamba blocks. The initial ``G" indicates ``guided". Please zoom in for a better view.}
    \label{fig: network}
\vspace{-1em}
\end{figure*}

\subsection{Atmospheric Turbulence Modeling}
Atmospheric turbulence simulation spans from computational optics \cite{Hardie_2017_a, Roggemann_2012_a, Roggemann_1995_a, Schmidt_2010_a} that rely on expensive wave computations to computer vision-oriented approaches \cite{Milanfar_2013_a, Lau_2019_a, Chak_2021_a} that offer speed but arguably lack physical foundations with others in the middle such brightness function-based simulations \cite{Vorontsov_2005_a, Lachinova_2007_a, Lachinova_2017_a} or learning-based alternatives \cite{Miller_2019_a, Miller_2021_a}, though speed remains a bottleneck for deep learning applications \cite{Mao_2021_a}. In this work, we utilize recent Zernike-based methods \cite{chimitt2022real, Mao_2021_a, Chimitt_2020_a, chan2023computational} as others previously \cite{Zhang_2022_a, Mao_2022_a, Jaiswal_2023_ICCV, jiang2023nert} due to their speed and ability to generalize to real-world sequences.

At the core of Zernike-based simulation is modeling the spatial varying point spread functions (PSF), which is the magnitude of the discrete Fourier transform (DFT) of the local \emph{phase distortion}, denoted by $\phi$. In the Zernike-based simulation framework, $\vphi = \sum_i a_i \mZ_i$ where $\mZ_i$ is the $i$th Zernike polynomial \cite{noll1976zernike} and $a_i$ is the $i$th Zernike coefficient sampled from a known distribution \cite{chan2023computational}. Zernike-based simulation takes a ground truth image $\boldsymbol{J} \in \R^{H\times W}$ and generates a random Gaussian vector field sample $\va \in \R^{H\times W}$ that describes the turbulence distortions \cite{chan2023computational}. Using numerically derived convolution kernels $\boldsymbol{\psi}_k$ as in \cite{Mao_2021_a, Chimitt_2023_b} for low-rank approximation, the simulation operates as function $g(\boldsymbol{J};\va)$:
\begin{equation}
    \boldsymbol{I} \bydef g(\boldsymbol{J};\va) = \sum_{k=1}^{100}\boldsymbol{\psi}_k \circledast (\boldsymbol{\beta}_{k} \cdot \warp(\boldsymbol{J}; \tilt)) + \boldsymbol{n},
    \label{eq: p2s_psf_basis}
\end{equation}
with $\warp$ is the spatial warp operation guided by pixel-shift field $\tilt$, $\boldsymbol{\beta}_{k}$ is the spatial-temporal weight of the corresponding $\boldsymbol{\psi}_k$, $\circledast$ denotes the depth-wise convolution, $\boldsymbol{I}$ is the turbulence degraded image and $\boldsymbol{n}$ is the white noise. $\tilt$ and $\boldsymbol{\beta}_{k}$ are functions of Zernike coefficient field $\va$ \cite{Mao_2021_a}.

\subsection{Turbulence Mitigation Methods}
Traditional turbulence mitigation (TM) algorithms, dating back to works like \cite{fraser1999atmospheric, Vorontsov_2001_a, Frakes_2001_a}, generally approach the problem as a many-to-one restoration task. They mostly follow a common pipeline, where the input frames are aligned, followed by an image fusion \cite{Milanfar_2013_a, Anantrasirichai_2013_a, Lou_2013_a, Mao_2020_a, Xie_2016_a, Oreifej_2013_a, Lau_2019_a, He_2016_a, xu2024longrange, Lao_2024_DTRATM}. For dynamic scenes with moving objects \cite{Gepshtein_2004_a, Oreifej_2013_a}, existing methods assume rigid motion in dynamic areas and rely on conventional pipelines for static regions \cite{Mao_2020_a, saha2024turb}. 

More deep-learning methods have achieved state-of-the-art turbulence mitigation results in recent years. Based on the input dimension, existing works can be categorized into single-frame based \cite{Lau_2021_a, Rai_2022_a, mei2023ltt, nair2023ddpm, Yasarla_2022_a, Nair_2021_a, Lau_2021_b, Jaiswal_2023_ICCV, Mao_2022_a, Xia_2024_NBGTR, Wang_2023_diff, wang2024real} and multi-frame based \cite{Jin_2021_a, Zhang_2022_a, Anantrasirichai_2022_a, lopez2023variational, zhang2024spatio} methods. While the single-frame TM models have a certain flexibility, the multi-frame ones are preferred when image sequences are available because they can leverage the additional information from an extended temporal respective field. Although the current CNN-based \cite{Jin_2021_a, Anantrasirichai_2022_a} approaches achieved temporal consistency in videos, they suffer from the limited temporal perceptive field. \cite{Zhang_2022_a} introduced temporal-channel self-attention to achieve longer-term information aggregation. However, the quadratic complexity hinders it from adapting to capturing very long temporal dependencies. The recurrent-based method \cite{zhang2024spatio} has a global temporal receptive field, while the nonlinear recurrent operation causes training inefficiency and inference instability.

\subsection{Selective State Space Models}
Recently, the selective state space models, represented by Mamba \cite{gu2023mamba, dao2024mamba2} have demonstrated efficiency in natural language modeling due to their linear scaling with sequence length in long-range dependency modeling. With this promising property, it exhibits great potential to be applied in multiple domains of computer vision \cite{visionmamba_2024, chen2024mim, li2024videomamba, oshima2024ssm, hu2024zigma, zhang2024surveymamba}. More recently, Mamba has been applied to various low-level vision tasks including general image restoration \cite{guo2024mambair}, image and video draining \cite{wu2024rainmamba, zou2024freqmamba}, image deblurring \cite{gao2024mambadeblur}, super-resolution \cite{lei2024dvmsr}, and low-light enhancement \cite{zou2024wave}, it has shown promising performance with relatively low cost than previous approaches. Since video turbulence mitigation requires a large receptive field, and the joint estimation of degradation patterns and clean images requires training and inference efficiency, we explore the potential of Mamba in turbulence mitigation and use it to serve as a strong baseline for future research.


\section{Method}
\label{sec:method}
\begin{figure*}[h]
    \centering
    \includegraphics[width = 0.98\linewidth]{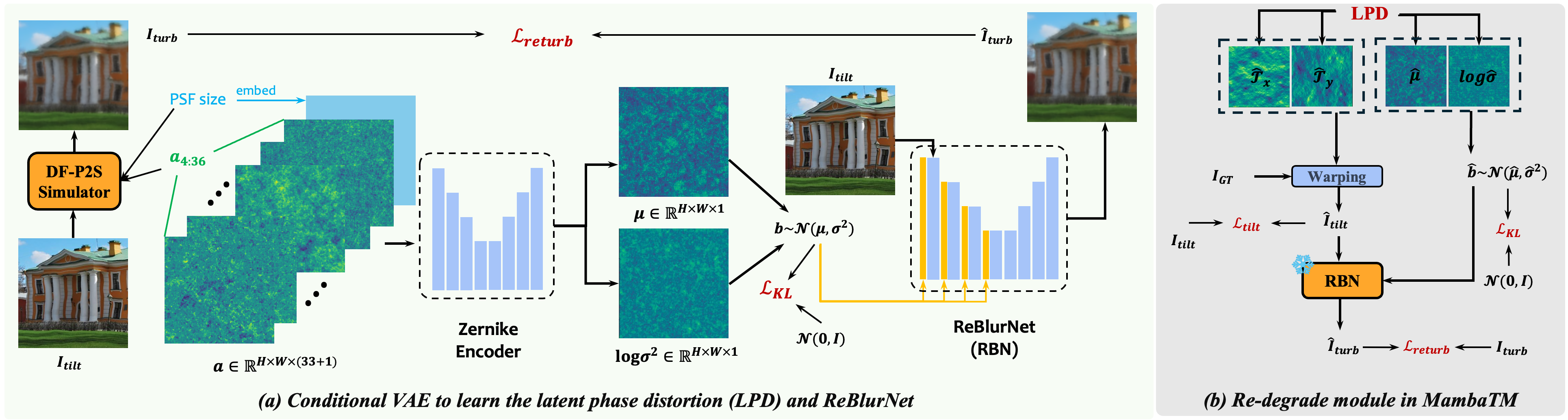}
    \vspace{-0.5em}
    \caption{The learning scheme of the latent phase distortion representation and ReBlur Network. Both the Zernike encoder and ReBlurNet are tiny NAFNet \cite{chen2022simple}, and the ReBlurNet's encoder part is modulated by the latent blur feature $\vb$. Please zoom in for a better view.}
    \label{fig: vae}
    \vspace{-0.5em}
\end{figure*}

\subsection{Overview}
Figure \ref{fig: network} shows the architecture of MambaTM for video turbulence mitigation. It consists of a multi-scale encoder, cascaded Mamba blocks, a latent phase distortion (LPD) encoder and decoder, and a multi-scale decoder, with all encoders and decoders operating on single frames. Given an image sequence $\boldsymbol{I} \in \mathbb{R}^{T\times H\times W\times 3}$, the encoder extracts multi-scale features, starting with spatial resolution $H \times W$ and channel $C$, halving the resolution and doubling the channels at each scale. The latent feature $\boldsymbol{x}$ is processed by $N_{1}$ Mamba groups, each containing three bidirectional Mamba blocks with different spatiotemporal scan orders. The LPD decoder estimates the phase distortion $D \in \mathbb{R}^{T\times H\times W\times 4}$, including tilt (first two channels) and blur (last two channels) representation. The LPD re-degrades the clean image via a variational module and guides subsequent $N_{2}$ Mamba groups. The LPD encoder compresses the LPD into $r_{LPD} \in \mathbb{R}^{T\times H/8\times W/8\times 8C}$ to help guide Mamba groups process $\boldsymbol{x}$ more effectively. These groups differ from previous ones by computing state space parameters based on both $\boldsymbol{x}$ and $r_{LPD}$. Finally, a multi-scale decoder is applied to produce the restored images $I_{restore}$.

The weights of the re-degrade module are frozen when training the MambaTM. It is trained before the restoration model via a conditional VAE (cVAE). The motivation and learning method will be introduced in the following section. 


\subsection{Learning Phase Distortion Representation}

Estimating and using the degradation representation as guidance is a common trick in image and video restoration \cite{Purohit_2021_ICCV, li2022allinone, li2022learning}. In our context, this requires the recovery of $\va$ from $\mI$. Although \autoref{eq: p2s_psf_basis} itself is differentiable and has been used in training previously \cite{feng2022TurbuGAN, Jaiswal_2023_ICCV}, the estimation of $\va$ from $\boldsymbol{I}$ (i.e., the Zernike estimation step) is ill-posed as we empirically show in Section \ref{sec:lpd_experiment}. The reason can be traced back to the phase retrieval problem \cite{sayre_1952_a, gerchberg_1972_a, fienup_1982_a}, though intuitively can be understood that there is an infinite number of solutions $\va' \neq \va$ that satisfy $g(\boldsymbol{J};\va) = g(\boldsymbol{J};\va')$. Furthermore, \autoref{eq: p2s_psf_basis} poses a computational bottleneck for co-training with an efficient multi-frame TM network.

To solve these two problems, we develop a parameterization trick using an efficient network and latent phase distortion (LPD) representation to replace the Zernike-based turbulence simulator, making the degradation estimation more well-posed and significantly faster to train while maintaining key physical properties.

\noindent \textbf{Latent phase distortion (LPD).} The proposed LPD methodology requires two primary components: a mapping that produces latent a representation $\widetilde{\va}$ from $\va$ and a reformulation of \autoref{eq: p2s_psf_basis} that utilizes $\widetilde{\va}$, i.e., $\widetilde{g}(\boldsymbol{J}, \widetilde{\va})$. The aim of the latent representation $\widetilde{\va}$ is to remove the ill-posedness related to multiple solutions and reduce dimensionality, while the modified degradation operator $\widetilde{g}$ uses $\widetilde{\va}$ directly and is significantly more efficient.

We use a variational autoencoder (VAE) to convert $\va$ to $\widetilde{\va}$, choosing $\widetilde{\va} \sim \calN(\vmu, \vsigma^2)$ where $\vmu$ and $\vsigma$ are the outputs of the VAE to be the form of the latent code. We refer to $\boldsymbol{\mu}$ and $\text{log}\boldsymbol{\sigma}$ as the \emph{Latent Phase Distortion (LPD)}, which has a one-to-one mapping to the actual turbulence degradation pattern, and illustrate the Zernike-to-LPD encoding in Figure \ref{fig: vae} which has the form of a UNet \cite{ronneberger2015u}.

To make use of the latent code $\widetilde{\va}$, we use a multi-scale reformulation of \autoref{eq: p2s_psf_basis} through a network we refer to as ReBlurNet (RBN) following the tilt-then-blur framework \cite{Chan_2022_a}. This multi-scale approaches saves significantly on computation, especially for large kernel sizes. The RBN first encodes the input image into multi-scale features, which are then modulated by multi-scale features of $\widetilde{\va}$ via element-wise product and decoded to get the blurred output image. More details about the modulation can be found in section 7.2 of the supplementary document. The encoder and decoder of the RBN have the same architecture as the Zernike Encoder. After training with the VAE, the weight of the RBN will be frozen and used in the re-degrade module in the MambaTM.

\noindent \textbf{Additional details.} It is relatively easy to predict the deformation field $\tilt$ with multiple frames \cite{Zhang_2022_a, Li_2021_a}. Therefore, we only reparameterize the higher-order Zernike coefficients to the latent variable $\vb$ representing the \emph{blur components}. Practically speaking, the LPD includes this deformation field $\tilt$ as a separate quantity. Furthermore, optical effects captured by the Zernike-based simulator depend on the kernel size (i.e., the spread of each $\vpsi_k$). To capture the dependence on the PSF size, we embed it as an additional channel to augment the Zernike coefficients. This gives us an additional benefit of having the kernel size being differentiable in estimation, which is crucial for applications in real scenes.

\subsection{Mamba blocks}
Estimating the degradation pattern and mitigating the turbulence jointly naturally requires more computation than solving the TM problem only. However, because the turbulence degradation can be viewed as a Gaussian random field \cite{chan2023computational}, multi-frame turbulence mitigation usually requires aggregating large chunks of frames to reliably estimate the underlying clean images \cite{zhang2024spatio}. Balancing the size of the spatiotemporal respective field and model efficiency is a major challenge for the video TM task. Existing methods \cite{Jin_2021_a, zhang2024spatio, Zhang_2022_a} try to expand the temporal perceptive field and rely on the convolution operation spatially. They all struggle to build long-range dependencies efficiently. Recently, sequence models with linear complexity, such as Mamba \cite{gu2023mamba} and RWKV \cite{peng2023rwkv}, have shown promise in obtaining a global receptive field for vision tasks \cite{guo2024mambair}, motivating us to explore their application in video turbulence mitigation.

\noindent \textbf{State Space Model (SSM).} Inspired by the classical state space model \cite{kalman1960new}, the SSM transformation \cite{gu_ssm} maps a 1D signal \(x(t) \in \mathbb{R}\) to a 1D output signal \(y(t) \in \mathbb{R}\) through an $N$-D latent state \(\boldsymbol{h}(t) \in \mathbb{C}^N\): $\boldsymbol{h}'(t) = A \boldsymbol{h}(t) + \boldsymbol{B} x(t)$, $y(t) = \boldsymbol{C} \boldsymbol{h}(t) + Dx(t)$, where \(\boldsymbol{A} \in \mathbb{C}^{N \times N}\) is the evolution parameter and \(\boldsymbol{B} \in \mathbb{C}^{N \times 1}, \boldsymbol{C} \in \mathbb{C}^{1 \times N}\), $D\in \mathbb{C}$ are controlling parameters. To process the discrete input sequence $\boldsymbol{x}=(x_0, x_1, ..., x_{L-1})\in \mathbb{R}^{L} $, following \cite{gupta2022diagonal}, Mamba \cite{gu2023mamba} employs the zero-order hold (ZOH) assumption to convert the continuous parameters \(\boldsymbol{A}, \boldsymbol{B}\) into their discrete counterparts \(\bar{\boldsymbol{A}}, \bar{\boldsymbol{B}}\). The calculation is shown in Figure \ref{fig: network}(c), and more details are provided in section 7.1.

\noindent \textbf{Phase distortion Guided SSM.} 
The re-degradation process has implicitly imposed degradation awareness on the model. However, the LPD map can also be used to explicitly guide the restoration. The original controlling parameters in the S6 model only depend on the input sequence $\boldsymbol{x}$ by the linear transforms $\boldsymbol{\Delta} = s_{\Delta}(\boldsymbol{x})$, $\boldsymbol{B}=s_{B}(\boldsymbol{x})$, and $\boldsymbol{C}=s_{C}(\boldsymbol{x})$. Changing these three parameters to LPD-dependent can effectively make input $\boldsymbol{x}$'s aggregation guided by the degradation information. We first encode the LPD to the same size as the image feature embeddings, then use the embedding of LPD $\boldsymbol{r}$ to modulate the input-dependent state parameters by $\boldsymbol{\Delta} = s_{\Delta}(\boldsymbol{x};\boldsymbol{r})$, $\boldsymbol{B}=s_{B}(\boldsymbol{x};\boldsymbol{r})$, and $\boldsymbol{C}=s_{C}(\boldsymbol{x};\boldsymbol{r})$. Figure \ref{fig: network} (c) shows the comparison between the original SSM and our phase distortion-guided SSM (GSSM). Moreover, LPD is also used to modulate the gating mechanism of the Mamba layer's output feature, as illustrated in Figure \ref{fig: network} (f). Our modification of the Mamba layer facilitates degradation-dependent state evolution and transition, improving the efficiency of SSM transformation.

\noindent \textbf{Mamba blocks.} 
When adapting Mamba to computer vision tasks, the 2D or 3D tensors are usually unfolded into 1D tokens \cite{zhang2024surveymamba}. The order of scanning, or flattening, impacts the model's performance. In vision tasks, the original Mamba's scanning design for 1D causal sequences is unsuitable for non-causal visual data. Therefore, we adopt the bidirectional scan \cite{visionmamba_2024} to address the characteristics of the video modality. Since different scan axis orders can result in different neighboring conditions, which result in different feature connectivity strengths along different axes, we applied three different scan orders. 
They are space-first scan, time-first scan, and local Hilbert scan \cite{wu2024rainmamba}. They are employed in Space-First Mamba Block (SFMB), Time-First Mamba Block (TFMB), and Local Hilbert Mamba Block (LHMB), respectively. The space-first scan follows a raster scan order to traverse along the \emph{width-height-time} axes order, and the time-first scan traverses along the order of the \emph{time-height-width} axes. Combining these two orders, we can obtain a relatively unbiased connectivity strength on three axes. However, the Mamba layer has a global perceptive field, and the 1D sequential model nature could cause weak connections on the neighboring pixels. The Hilbert scan is then used to address this. The Hilbert curve \cite{hilbert1935stetige} is a space-filling curve that addresses preserving locality \cite{jagadish1990linear} when flattening multi-dimensional data. It is designed to optimally enforce the elements close to each other in the multi-dimensional space to remain closed when flattened to 1D and has shown effectiveness when used in the video draining task \cite{wu2024rainmamba}. Therefore, we add LHMB as the third Mamba block in each Mamba group.

\subsection{Losses}
Our training has two stages: the ReBlurNet training and MambaTM training. The ReBlurNet training follows the typical VAE framework, where the turbulence re degradation loss $\mathcal{L}_{returb}$ is the L1 loss computed between the DF-P2S \cite{chimitt2022real, zhang2024spatio} simulated images $I_{turb}$ and the ReBlurNet output images $\hat{I}_{turb}$, the KL divergence loss $\mathcal{L}(\boldsymbol{\mu}, \boldsymbol{\sigma}^2)$ is used to enforce the sampled latent blur representation $\boldsymbol{h}$ to be close to the Gaussian distribution $\mathcal{N}(\boldsymbol{0}, \boldsymbol{I})$:
\begin{equation}
    \mathcal{L}_{KL} = -\frac{0.5}{H\times W}\sum_{i,j}(\text{log}(\boldsymbol{\sigma}_{i,j}^2)+1-\boldsymbol{\mu}_{i,j}-\boldsymbol{\sigma}_{i,j})
\end{equation}
Finally, the ReBlurNet training has the VAE loss: 
\begin{equation}
\mathcal{L}_{VAE} = \mathcal{L}_{returb} + \alpha_{k}\mathcal{L}_{KL}
\end{equation}

\noindent In the MambaTM training, the ReBlurNet is fixed, and we jointly optimize the turbulence re-degradation and mitigation. The overall loss is computed as a combination of the restoration loss $\mathcal{L}_{restore}$ and the re-degradation loss $\mathcal{L}_{returb}$. The restoration loss is denoted as:
\begin{equation}
    \mathcal{L}_{restore}(\hat{\boldsymbol{J}}, \boldsymbol{J}) = \mathcal{L}_{c}(\hat{\boldsymbol{J}}, \boldsymbol{J}) + \alpha_{p} \mathcal{L}_{p}(\hat{\boldsymbol{J}}, \boldsymbol{J})
\end{equation}
Where $\mathcal{L}_{c}$ is the Charbonnier loss \cite{charbonnier} and $\mathcal{L}_{p}$ is the perceptual loss \cite{lpips}, $\alpha_{p}$ is the weight for the perceptual loss.
\noindent On the other hand, the re-degradation loss is computed by:
\begin{equation}
    \mathcal{L}_{returb} = \mathcal{L}_{c}(\hat{\boldsymbol{I}}_{tilt}, \boldsymbol{I}_{tilt}) + \mathcal{L}_{c}(\hat{\boldsymbol{I}}_{turb}, \boldsymbol{I}) + \alpha_{k}\mathcal{L}_{KL}
\end{equation}
where the $\boldsymbol{\hat{I}}_{tilt}$ is the deformed image warped by the estimated tilt $\hat{\tilt}$, $\boldsymbol{\hat{I}}_{turb}$ is the re-degraded image and $\boldsymbol{I}$ is the degraded image in equation \ref{eq: p2s_psf_basis}. The overall loss for the MambaTM training is 
\begin{equation}
    \mathcal{L} = \mathcal{L}_{restore} + \alpha \mathcal{L}_{returb}
\end{equation} 
We empirically set $\alpha = 0.2$, $\alpha_{p} = 0.01$ and $\alpha_{k}=0.001$.
\section{Experiment}
\label{sec:experiment}
\subsection{The LPD or Zernike Representation}
\label{sec:lpd_experiment}

\begin{table}
\centering
\resizebox{0.47\textwidth}{!}{
\begin{tabular}{c|ccc}
\hline
Representation & Speed (s) & $\text{PSNR}_{returb}$ & Differentiable \\
\hline
Zernike &  $0.16 \sim 6.10$  & 25.84 / 31.17  & Partial \\
LPD [Ours]  &  0.02  &  34.08 & Full \\
\hline
\end{tabular}
}
\caption{Comparison of different phase distortion representations. The variance of the speed in the Zernike-based simulator \cite{chimitt2022real, chan2023computational, zhang2024spatio} is caused by different blur kernel sizes. Two values of Zernike-based representation's PSNR are the re-degradation performance under rigid supervision (left) and loose supervision (right). }
\vspace{-0.5em}
\label{table:lpd}
\end{table}

\begin{table*}
\centering
\setlength{\aboverulesep}{0pt}
\setlength{\belowrulesep}{0pt}
\resizebox{0.99\textwidth}{!}{
\begin{tabular}{l|ccc|c}
\toprule[1pt]
    Turbulence Level & Weak & Medium & Strong  & Overall \\
    \hline
Methods & PSNR / SSIM / LPIPS & PSNR / SSIM / LPIPS & PSNR / SSIM / LPIPS & PSNR / SSIM / LPIPS \\
\hline
RNN-MBP \cite{zhu2022deep} & 27.9243 / 0.8438 / 0.2096 & 27.4742 / 0.8210 / 0.2178 & 26.0812 / 0.7900 / 0.2511 & 27.2161 / 0.8186 / 0.2245 \\
ESTRNN \cite{zhong2022real} & 28.9805 / 0.8622 / 0.2005 & 28.3338 / 0.8472 / 0.2063 & 26.8897 / 0.8076 / 0.2480 & 28.1347 / 0.8407 / 0.2169 \\
VRT \cite{liang2022vrt} & 28.8453 / 0.8625 / 0.1831 & 28.2628 / 0.8492 / 0.1865  & 26.7492 / 0.8217 / 0.2207 & 28.0179 /  0.8442 / 0.1954 \\
RVRT \cite{liang2022rvrt} & 29.8950 / 0.8799 / 0.1806 & 29.1658 / 0.8686 / 0.1855 & 27.6827 / 0.8309 / 0.2221 & 28.9332 / 0.8656 / 0.1957 \\
\hline
TSRWGAN \cite{Jin_2021_a} & 27.0844 / 0.8435 / 0.2141 & 26.7046 / 0.7915 / 0.2221 & 25.4230 / 0.7358 / 0.2671 &  26.4541 / 0.7927 / 0.2325  \\
TMT \cite{Zhang_2022_a} & 29.1183 / 0.8654 / 0.1820  & 28.5050 / 0.8524 / 0.1841  & 26.9744 / 0.8110 / 0.2206 &  28.2665 / 0.8430  / 0.1942 \\
DATUM \cite{zhang2024spatio} & 30.2058 / 0.8867 / 0.1788 & 29.6203 / 0.8783 / 0.1825 & 28.2550 / 0.8456 / 0.2188 & 29.4222 / 0.8714 / 0.1919 \\
\rowcolor{Gray}MambaTM [Ours] & \textbf{30.8736 / 0.8991 / 0.1425} & \textbf{30.0816 / 0.8903 / 0.1426} & \textbf{28.6142 / 0.8601 / 0.1721} & \textbf{29.9151 / 0.8843 / 0.1516} \\
\bottomrule[1pt]
\end{tabular}
}
\vspace{-0.5em}
\caption{Performance comparison on the ATSyn-dynamic set \cite{zhang2024spatio}, we list the image quality scores on different turbulence levels.}
\vspace{-0.5em}
\label{table:dynamic}
\end{table*}


We conduct experiments to demonstrate the necessity of learning a latent representation of the phase distortion instead of the classical Zernike-based degradation representation by evaluating re-degrade performance. Specifically, we first pre-train a MambaTM network for restoration only and then modify the output modality to either the Zernike coefficient map or the LPD coefficient map. The former is a 35-channel tensor (two for tilt and 33 for blur) while the latter is also a 35-channel tensor, both maintaining the same spatial-temporal dimensions as the input images. Additionally, the Zernike-based simulator \cite{chimitt2022real, chan2023computational, zhang2024spatio} requires the PSF size, for which we add a regression head on the output with the Sigmoid function and linearly scale it to odd values between 3 and 99. For Zernike coefficient estimation, we explored both rigid supervision, which solely uses the ground truth Zernike random field as the supervision signal, and loose supervision, which utilizes the degraded images as the supervision signal. We finetune the pre-trained MambaTM for $100,000$ iterations with batch size 1. Table \ref{table:lpd} presents the re-degradation results along with two other practical factors: speed and differentiability.

The table shows that learning Zernike coefficients presents significant challenges for supervised training, as numerous solution combinations can produce identical turbulence profiles. When we apply supervision to the Zernike coefficients, the training cannot even converge. When we apply supervision on images, the ill-posedness is alleviated, and the model can converge to local minima. In contrast, predicting LPD coefficients enables the model to deliver substantially improved re-degradation performance. Additionally, the LPD-based simulation is more straightforward as it eliminates the need for a regression head to determine PSF size while operating much faster than the Zernike-based simulator. 

\subsection{Datasets and Training Scheme}
We trained the conditional ReBlurNet with the VAE in a frame-wise manner. The ground truth Zernike random fields are generated on the fly with the data synthesis method introduced in \cite{zhang2024spatio}. All turbulence conditions are randomly sampled from the condition parameters in the training set of the ATSyn dataset \cite{zhang2024spatio} and evaluated with the conditions in the test set of ATSyn. The clean images are sampled from the LSDIR dataset \cite{Li_2023_lsdir}. We set batch size 32 and trained the VAE for $10,000$ iterations. We use the Adam optimizer \cite{kingma2014adam} with the Cosine Annealing learning rate schedule \cite{loshchilov2016sgdr}, the learning rate is decayed from $0.001$ to $10^{-6}$. The reconstruction loss gets 46.5 dB PSNR on the test set with different turbulence conditions and image content from the training set. This indicates that our CRBN can reproduce the turbulence effect accurately and that the LPD representation is robust.

The MambaTM is trained and evaluated on the ATSyn dataset \cite{zhang2024spatio}. We trained two models for the dynamic scene and static scene modality separately. We first train our model on the ATSyn-dynamic set for $1.2\times 10^{6}$ iterations in a progressive training way. Specifically, we set the batch size 16, patch size $192\times 192$, and 18 frames at the beginning of training and gradually enlarged the input dimension and reduced the batch size. Finally, we changed the setting to batch size 4, 36 frames, and patch size $256\times 256$. We use the Adam optimizer with the Cosine Annealing learning rate scheduler, and the learning rate decays from $0.0002$ to $10^{-7}$. Consequently, we finetune our model on the ATSyn-static dataset for $6\times 10^{5}$ iterations to adapt to the static scene scenario. The entire training is conducted on 2 NVIDIA A100 GPUs with PyTorch implementation.

\subsection{Quantitative Comparison}
\noindent \textbf{On dynamic scene.} We demonstrate MambaTM's advantage quantitatively on ATSyn \cite{zhang2024spatio} and TMT's synthetic dataset \cite{Zhang_2022_a}. Following \cite{zhang2024spatio}, we also compare our methods with general video restoration networks \cite{liang2022vrt, liang2022rvrt,zhong2022real,zhu2022deep} and video TM networks  \cite{Jin_2021_a, Zhang_2022_a, Mao_2022_a, zhang2024spatio}. Except for \cite{zhang2024spatio}, which provides the trained model on the same benchmarks, we re-trained them under the training settings listed in the original papers. 
The comparison on the ATSyn-dynamic dataset is shown in Table \ref{table:dynamic}. We compare our model against the others on the pixel-wise score PSNR, SSIM, and perceptual score LPIPS \cite{lpips}. The result indicates the clear advantage of our MambaTM in terms of reconstruction quality.

Besides the ATSyn-dynamic dataset, we also conduct a comparison on TMT's synthetic dataset \cite{Zhang_2022_a}. As shown in Table \ref{table:TMT_dynamic}, our MambaTM achieves SOTA performance on this benchmark as well. It is worth noting that our method's speed measuring with frame-per-second (FPS) is almost double that of the previous SOTA \cite{zhang2024spatio} and reaches real-time restoration on RTX 3090 GPU. The model size and MACs can be found in section 8 of the supplementary document.

\begin{table}
\centering
\small
\setlength{\aboverulesep}{0pt}
\setlength{\belowrulesep}{0pt}
\resizebox{0.48\textwidth}{!}{
\begin{tabular}{lccc|c}
\toprule[1pt]
 Methods & PSNR & SSIM & LPIPS & FPS \\
\midrule
VRT \cite{liang2022vrt} & 27.6114 & 0.8300 & 0.2485 & 0.38 \\
RVRT \cite{liang2022rvrt} & 27.8512 & 0.8388 & 0.2260 & 7.92 \\
TSRWGAN \cite{Jin_2021_a} & 26.3262 & 0.7957 & 0.2606 & 1.67 \\ 
TMT \cite{Zhang_2022_a} & 27.7419 & 0.8318 & 0.2475 & 1.50 \\
DATUM \cite{zhang2024spatio}  & 28.6006 & 0.8441 & 0.2245 & 32.7\\
\rowcolor{Gray} MambaTM [ours]  & \textbf{28.9049} & \textbf{0.8561} & \textbf{0.1996} & \textbf{55.4}\\
\bottomrule[1pt]
\end{tabular}
}
\vspace{-0.5em}
\caption{Performance on the TMT \cite{Zhang_2022_a}'s dynamic scene dataset and the speed of all TM and general restoration networks. The frame-per-second (FPS) is measured on $512\times 512$ patches on NVIDIA A100 GPU.} 
\label{table:TMT_dynamic}
\end{table}

\noindent \textbf{On static scene.} Next, we compare our method with others on the static scene scenarios. As shown in Table \ref{table:static}, our method reached the SOTA performance on the synthetic dataset ATSyn-static \cite{zhang2024spatio} in terms of PSNR and SSIM. On the real-world benchmark, evaluating the performance with pixel-wise similarity is impossible due to the absence of ground truth images. Real-world comparison usually involves comparing with pseudo ground truth \cite{Mao_2022_a}, or restored images to downstream tasks \cite{saha2024turb, Jaiswal_2023_ICCV, zhang2024spatio, qin2025unsupervised}. \cite{UG2} provides the Turb-Text dataset, where three popular text recognition models CRNN \cite{shi2016end}, DAN \cite{wang2020decoupled}, and ASTER \cite{shi2018aster} are applied to the restored images. A higher recognition rate can indicate higher-quality images. Benchmarking our model trained on the ATSyn-static with the Turb-Text dataset, we find that our method successfully restores text patterns under most turbulence conditions. Three models on the recovered images get over 98\% text recognition rate, reaching a new SOTA on the Turb-text dataset.

\begin{table}
\centering
\small
\setlength{\aboverulesep}{0pt}
\setlength{\belowrulesep}{0pt}
\resizebox{0.48\textwidth}{!}{
\begin{tabular}{lccc}
\toprule[1pt]
Benchmark & \multicolumn{2}{c}{ATSyn-static \cite{zhang2024spatio}} & Turb-Text ($\%$) \\
 \cmidrule(r){1-1} \cmidrule(r){2-3} \cmidrule(r){4-4} 
 Methods & PSNR & SSIM & CRNN/DAN/ASTER \\
\midrule
VRT \cite{liang2022vrt} & 24.2776 & 0.7180 & 76.30 / 84.45 / 83.60 \\
RVRT \cite{liang2022rvrt} & 25.7702 & 0.7415 & 86.40 / 89.00 / 89.20 \\
ESTRNN \cite{zhong2022real} & 26.3251 & 0.7760 & 87.10 / 97.80 / 96.95 \\
TSRWGAN \cite{Jin_2021_a} & 23.2291 & 0.6662 & 60.30 / 73.90 / 74.40 \\ 
TMT \cite{Zhang_2022_a} & 24.5112 & 0.7184 & 80.90 / 87.25 / 88.55 \\
DATUM \cite{zhang2024spatio}  & 26.7623 & 0.7817 & 93.55 / 97.95 / 97.25 \\
\rowcolor{Gray} MambaTM [ours]  & \textbf{27.0082} & \textbf{0.8044} & \textbf{97.80 / 99.35 / 98.15} \\
\bottomrule[1pt]
\end{tabular}
}
\vspace{-0.5em}
\caption{Static scene modality. CRNN/DAN/ASTER are the text recognition rates of these three models from the restored images.}
\label{table:static}
\vspace{-0.5em}
\end{table}

\subsection{Real-world Qualitative Comparison}
\begin{figure*}[ht]
    \captionsetup[subfloat]{font=scriptsize}
    \centering
  \subfloat[Input frame]{%
      \includegraphics[width=0.14\linewidth]{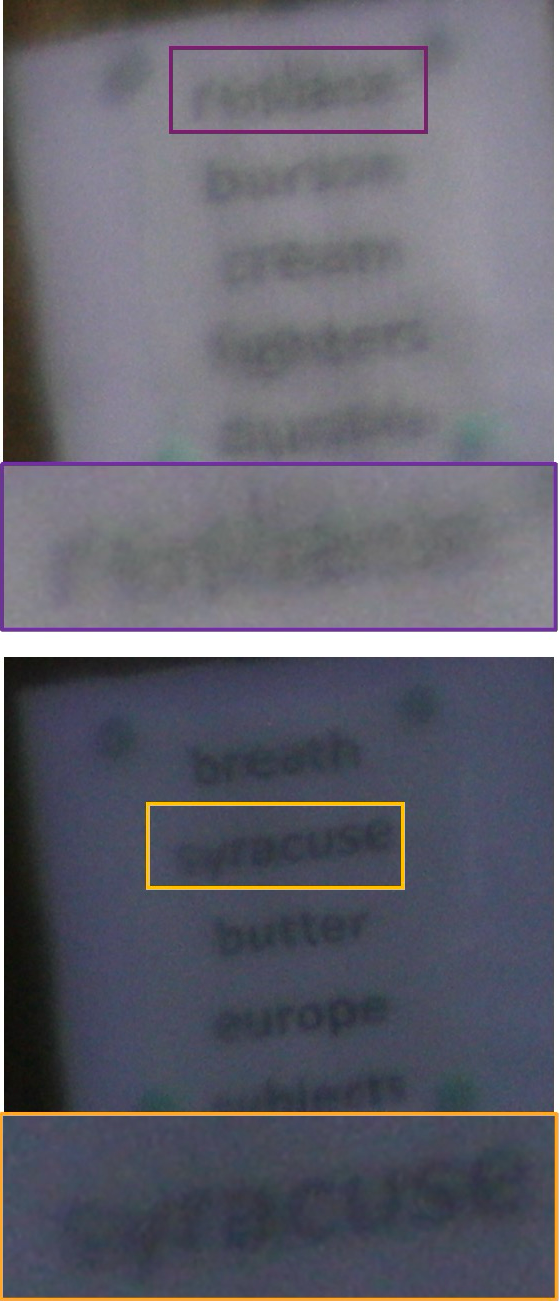}}
    \hfill
 \subfloat[PiRN \cite{Jaiswal_2023_ICCV}]{%
    \includegraphics[width=0.14\linewidth]{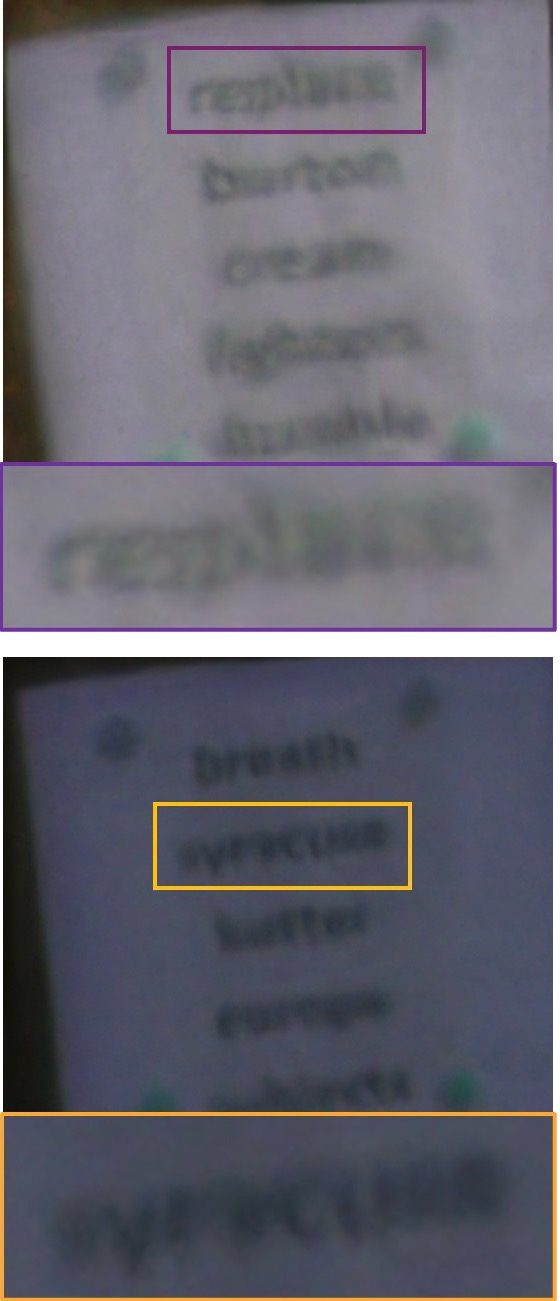}}
   \hfill
 \subfloat[TSRWGAN \cite{Jin_2021_a}]{%
      \includegraphics[width=0.14\linewidth]{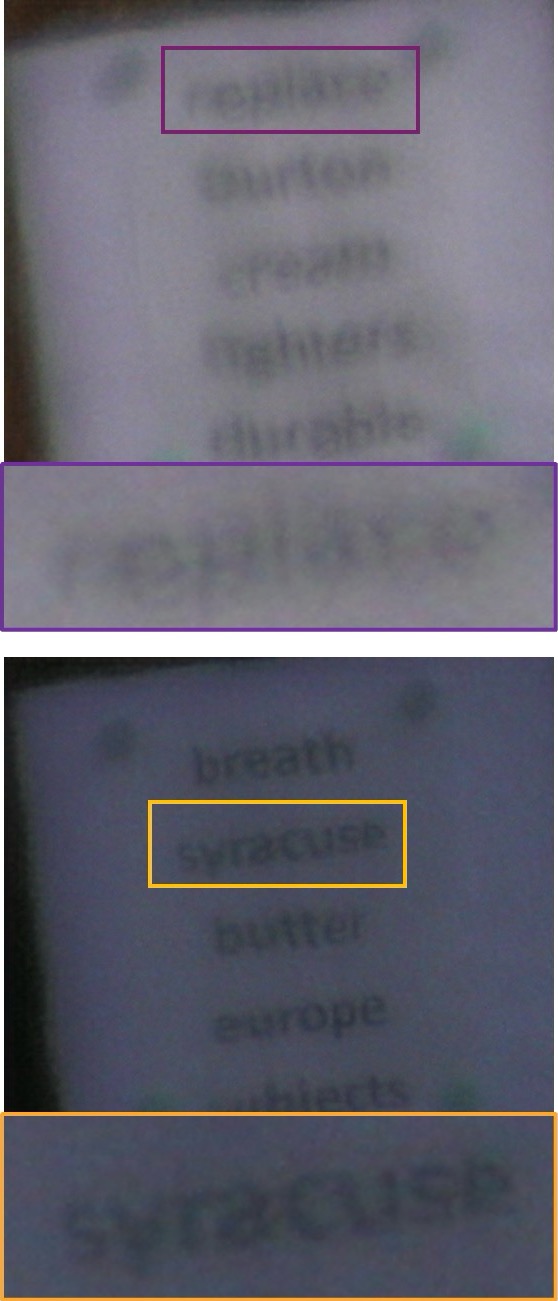}}
    \hfill
  \subfloat[TMT \cite{Zhang_2022_a}]{%
        \includegraphics[width=0.14\linewidth]{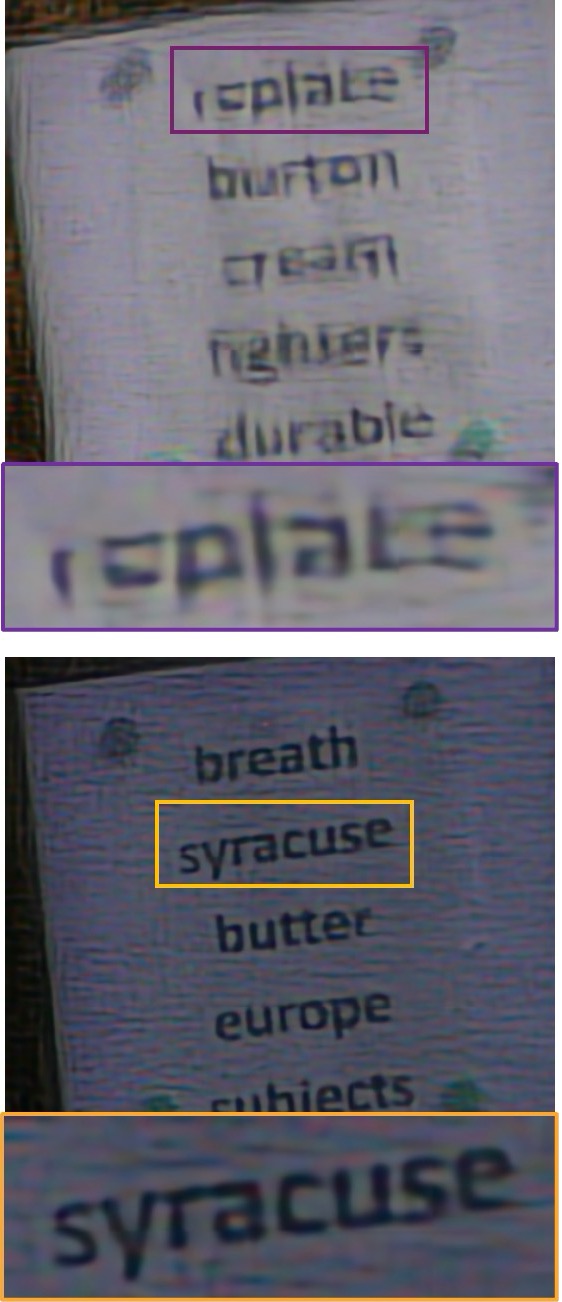}}
    \hfill
  \subfloat[Turb-Seg-Res \cite{saha2024turb}]{%
      \includegraphics[width=0.14\linewidth]{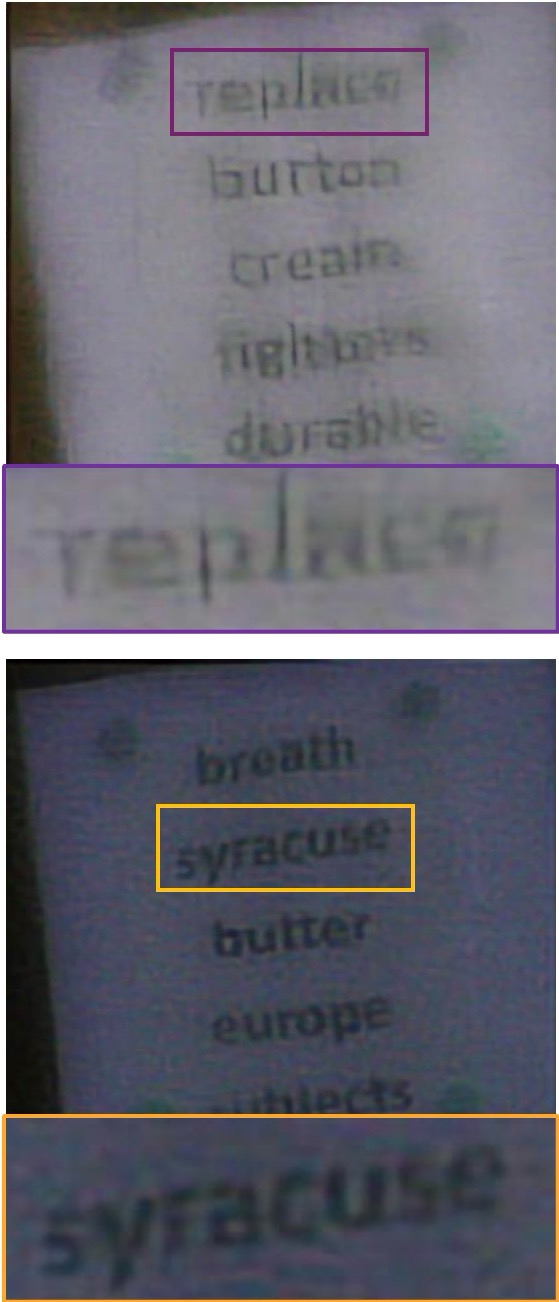}}
    \hfill
  \subfloat[DATUM \cite{zhang2024spatio}]{%
        \includegraphics[width=0.14\linewidth]{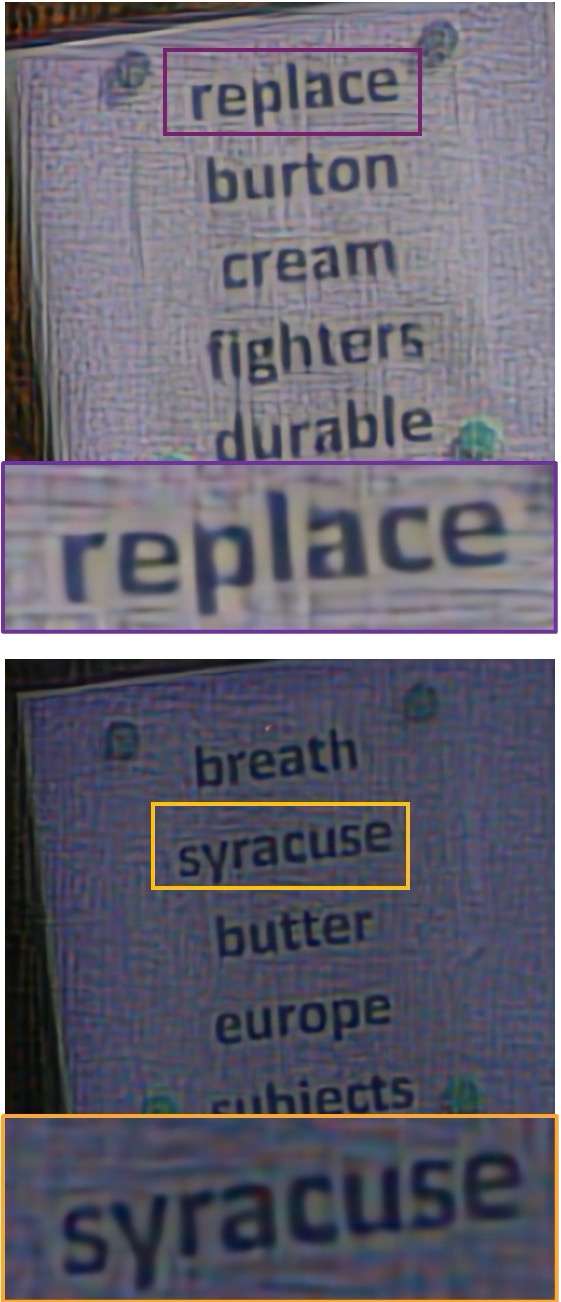}}
    \hfill
  \subfloat[MambaTM]{%
      \includegraphics[width=0.14\linewidth]{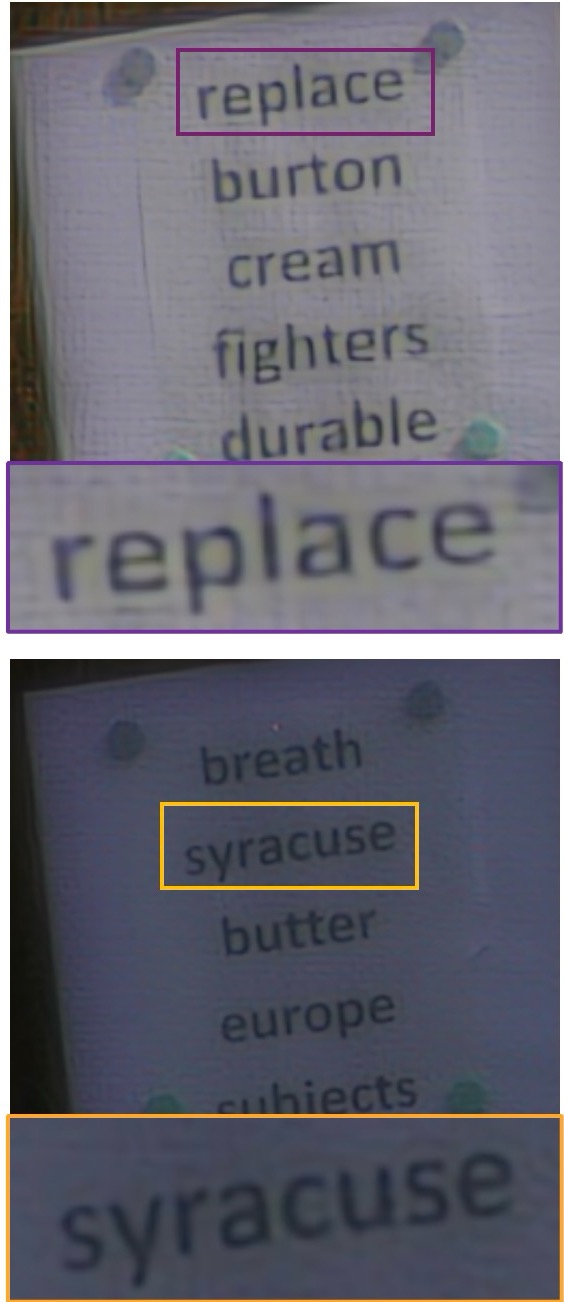}}
    \hfill
\vspace{-0.5em}
  \caption{Qualitative comparison on the turbulence-text dataset \cite{UG2}. The input frames (a) are from the 1st and 41st sequences in \cite{UG2}. Note that although (f) has a stronger contrast, it still contains much more color noise, which can be observed more clearly by zooming in.}
  \label{fig:text} 
\end{figure*}
\begin{figure*}[ht]
    \captionsetup[subfloat]{font=scriptsize}
    \centering
  \subfloat[Input frame]{%
      \includegraphics[width=0.163\linewidth]{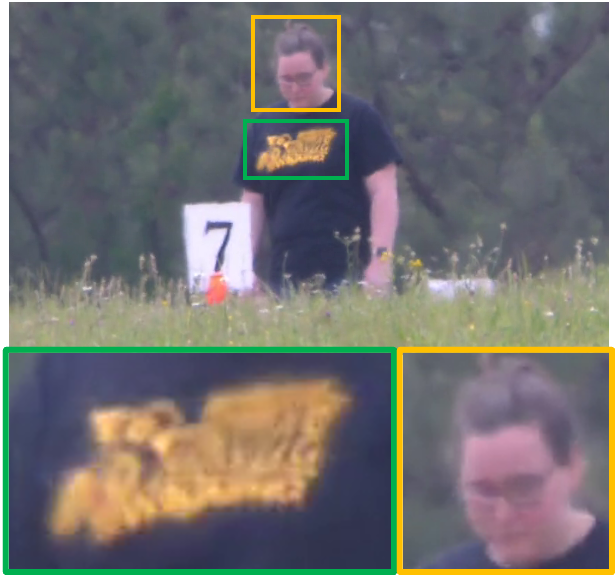}}
    \hfill
 \subfloat[TSRWGAN \cite{Jin_2021_a}]{%
      \includegraphics[width=0.163\linewidth]{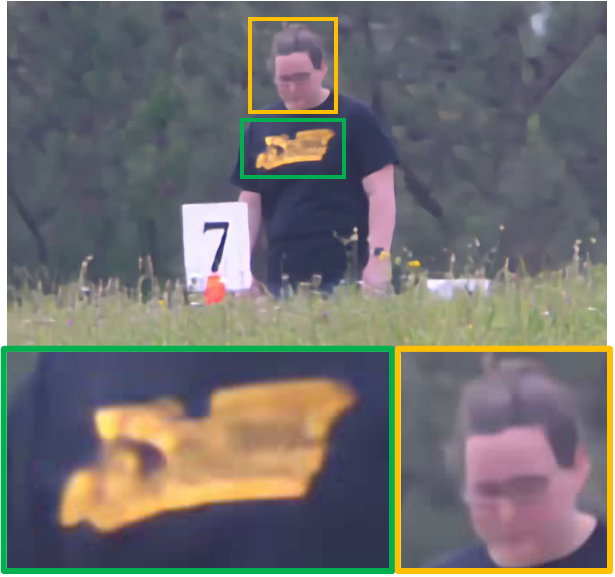}}
    \hfill
  \subfloat[TMT \cite{Zhang_2022_a}]{%
        \includegraphics[width=0.163\linewidth]{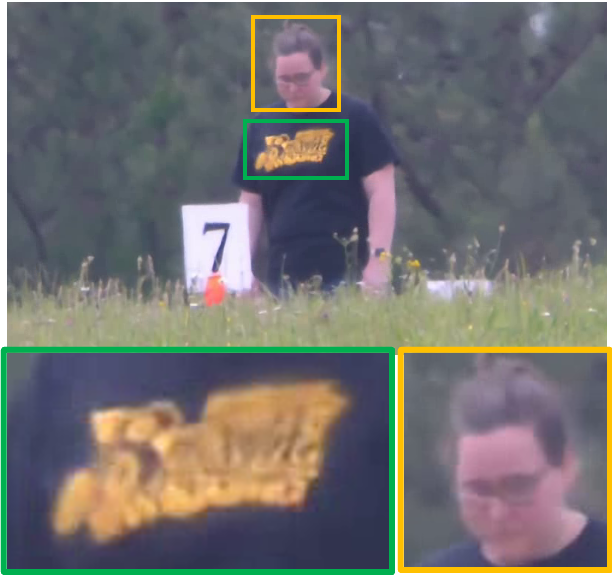}}
    \hfill
  \subfloat[Turb-Seg-Res \cite{saha2024turb}]{%
      \includegraphics[width=0.163\linewidth]{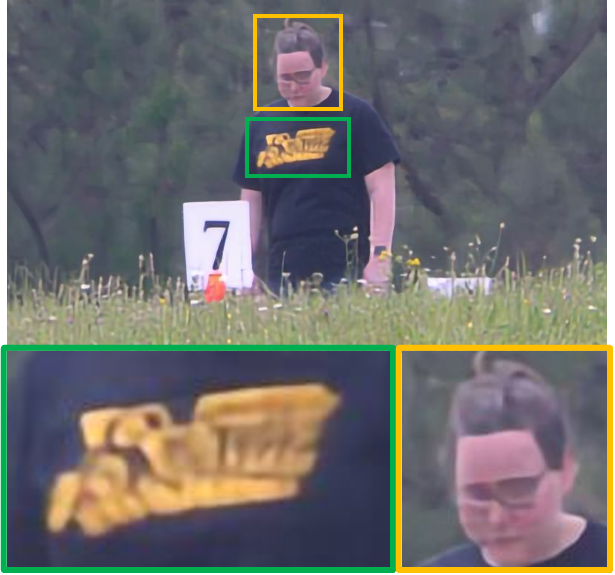}}
    \hfill
  \subfloat[DATUM \cite{zhang2024spatio}]{%
        \includegraphics[width=0.163\linewidth]{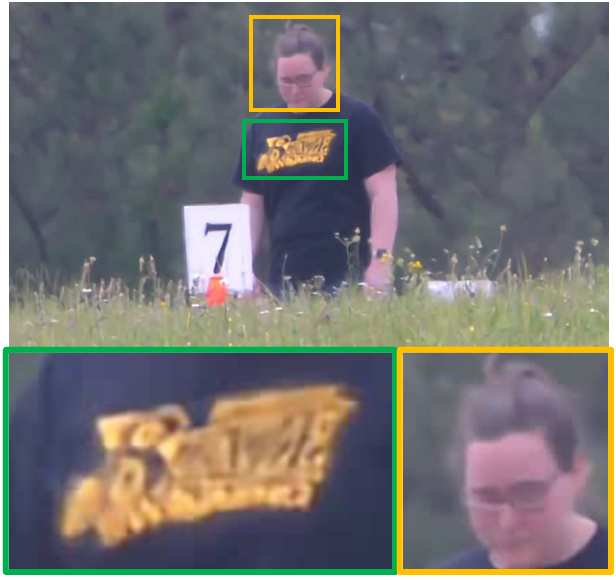}}
    \hfill
  \subfloat[MambaTM]{%
      \includegraphics[width=0.163\linewidth]{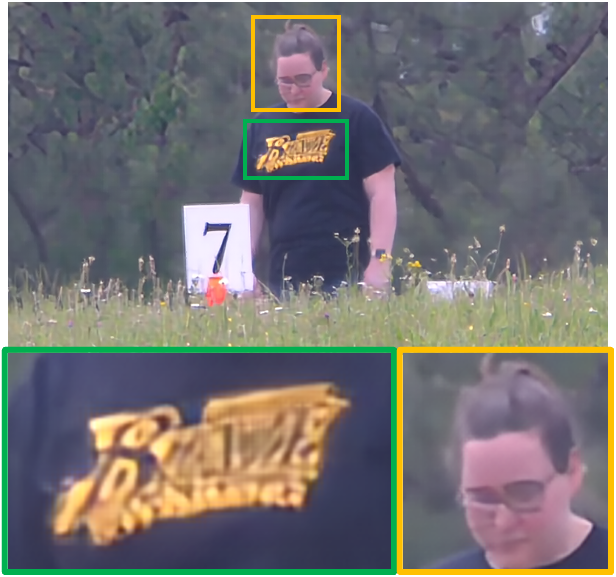}}
    \hfill
\vspace{-0.5em}
  \caption{Qualitative comparison on the BRIAR dataset \cite{cornett2023expanding}. The subject has given consent to publish the image. }
  \label{fig:briar} 
\end{figure*}

We also offer comparisons of real-world static and dynamic scene videos to demonstrate the advancements of our model. Figure \ref{fig:text} shows the results of different TM models on the Turb-text dataset. We can observe that our model reveals a clean text pattern under heavy turbulence. It is worth noting that despite being trained on the same datasets and similar settings, images processed by DATUM \cite{zhang2024spatio}, the recent SOTA have substantially more artifacts than MambaTM's images. This issue stems from the out-of-distribution noise and the instability of its non-linear recurrent operation. The SSM, a linear recurrent model, can effectively alleviate the instability problem. Besides the static scene images, a comparison of the real-world dynamic scene images is provided in Figure \ref{fig:briar}, from which we can observe that our model can recover more details than other models from a degraded image. More visual comparisons are provided in the supplementary material, since the turbulence degradation is temporally varying, we highly recommend readers to watch our video samples.

\subsection{Ablation Study}
In this section, we conduct controlled experiments and compare our key design elements with alternatives to demonstrate the effectiveness of our specific design decisions.
\noindent \textbf{The design of RBN.} Since the RBN is employed in the re-degradation module, it must be accurate from the outset, meaning it should align with the images generated by the Zernike-based simulator. Surprisingly, a plain UNet achieves over 40 dB PSNR in re-degradation performance. We then replaced the UNet with the more advanced NAFNet \cite{chen2022simple}, resulting in a significant improvement. Additionally, we explored the effectiveness of incorporating a multi-scale modulator into the RBN encoder. With this modulator, we observed an improvement of over 1 dB. Following the original NAFNet design, the network size is controlled by adjusting the depth of the third-scale encoder layer and the number of channels. We modified these parameters to balance computational cost and performance. All design choices are listed in Table \ref{table:RBN}. 

\begin{table}
\centering
\resizebox{0.45\textwidth}{!}{
\begin{tabular}{l|ccc}
\hline
Models & PSNR & Speed (s) & Memory (GB) \\
\hline
PlainUNet* & 42.08 & 0.72 & 24.7 \\
PlainUNet &  43.26  &  0.82 & 26.3 \\
\hline
NAFNet*-8,16  & 44.93  &  0.62 & 55.9 \\
\rowcolor{Gray} NAFNet-8,16 & 46.72  & 0.70 & 57.5 \\
NAFNet-12,16 &  46.94 & 0.80 &  68.3 \\
NAFNet-8,8 &  42.89  & 0.49  & 32.1 \\
\hline
\end{tabular}
}
\vspace{-0.3em}
\caption{Comparison of different architecture choices of the RBN network. The consumption is measured with batch size 32. The $*$ indicates only modulating the encoder features on the first scale; others utilize the multi-scale modulation strategy. The numbers after NAFNet indicate the depth of the third scale encoder and the width of the network. Our final choice is marked in gray.}
\vspace{-0.5em}
\label{table:RBN}
\end{table}

\noindent \textbf{The scan mechanism.} We further investigate the impact of different scan strategies in MambaTM. We do this by removing one scan order each time and keeping the total number of SSM transformations the same by repeating other scan orders more times. To improve the experimental efficiency, we didn't incorporate the LPD guidance. The experiment result is listed in Table \ref{table:scan}, where we can find that the SSM is quite robust and adaptive to different scan orders; removing any component will not cause a dramatic performance drop. Despite this, hybridizing different scan orders is still beneficial, providing more homogeneous connectivity along different spatiotemporal axes to the model.

\begin{table}
\centering
\small
\setlength{\aboverulesep}{0pt}
\setlength{\belowrulesep}{0pt}
\begin{tabular}{cccccccc}
\toprule[0.8pt]
BD & SF & TF & LH & LPD & PSNR  & SSIM &  LPIPS\\
\midrule
\cmark & \cmark &  & \cmark  &   & 29.1274 & 0.8720  & 0.1671 \\
\cmark & & \cmark & \cmark  &   & 29.5058 & 0.8797 & 0.1570  \\
\cmark & \cmark & \cmark &  &  & 29.6077 & 0.8822 & 0.1565  \\
 & \cmark & \cmark & \cmark &  & 29.4933 & 0.8812 & 0.1594 \\ 
\cmark & \cmark & \cmark & \cmark & &29.6679 & 0.8830 & 0.1568  \\
\midrule
\cmark & \cmark & \cmark & \cmark & \cmark  & 29.7495 & 0.8808 & 0.1533  \\
\rowcolor{Gray} \cmark & \cmark & \cmark & \cmark & \cmark \cmark & 29.9151 & 0.8843 & 0.1516  \\
\bottomrule[0.8pt]
\end{tabular}
\vspace{-0.5em}
\caption{Ablation study on the MambaTM design. We tested the effectiveness of different scan directions. BD denotes bi-directional scan, SF, TF, and LH denote spatial-first, time-first, and local Hilbert scan, respectively. For the LPD, single \cmark means it is only used for reproducing the turbulence degradation; double \cmark means the LPD-guided Mamba block is also equipped.}
\vspace{-0.8em}
\label{table:scan}
\end{table}

\noindent \textbf{The LPD guidance.} The LPD guides MambaTM in two key ways: 1) It provides additional re-degradation supervision, allowing the turbulence properties embedded in the RBN to be implicitly transferred into MambaTM through backpropagation. 2) It facilitates degradation-aware state space construction, explicitly guiding MambaTM via the Guided Mamba layers. We evaluate these two aspects separately. First, when the re-degradation module is enabled and the Guided Mamba Layers are replaced with standard Mamba Layers, we observe certain improvements. Moreover, the Guided Mamba Layers offer a further performance boost, as shown in Table \ref{table:scan}. This experiment demonstrates the effectiveness of learning latent phase distortion for mitigating turbulence. More visualization of the LPD can be found in section 9 of the supplementary material.

\begin{figure*}[ht]
    \captionsetup[subfloat]{font=scriptsize}
    \centering
  \subfloat[input frame]{%
      \includegraphics[width=0.163\linewidth]{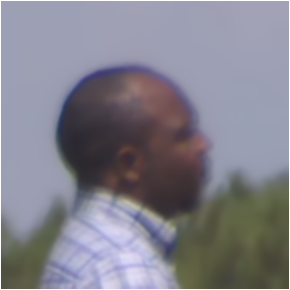}}
    \hfill
 \subfloat[restored]{%
    \includegraphics[width=0.163\linewidth]{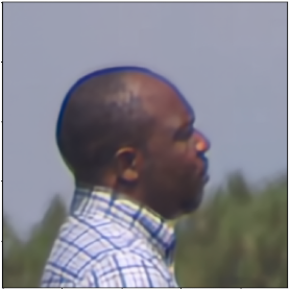}}
   \hfill
 \subfloat[re-degraded]{%
      \includegraphics[width=0.163\linewidth]{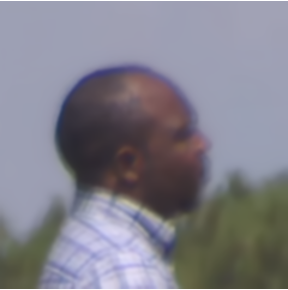}}
    \hfill
  \subfloat[$\boldsymbol{\mu}$]{%
        \includegraphics[width=0.163\linewidth]{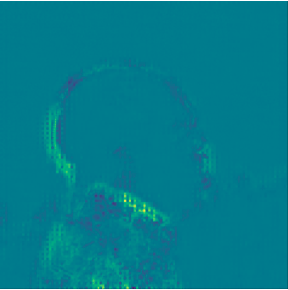}}
    \hfill
  \subfloat[$\boldsymbol{\sigma}$]{%
      \includegraphics[width=0.163\linewidth]{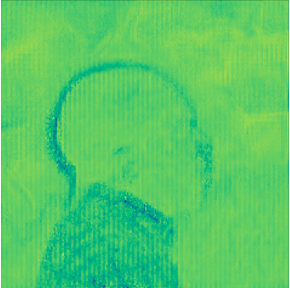}}
    \hfill
  \subfloat[$\widetilde{\va}$]{%
        \includegraphics[width=0.163\linewidth]{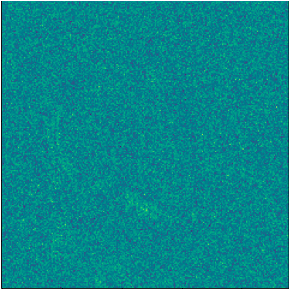}}
    \hfill
\vspace{-0.5em}
  \caption{Example of the blur representation in a real-world case from the BRIAR dataset \cite{cornett2023expanding}. We can view $\boldsymbol{\mu}$ as the blur strength bias, $\boldsymbol{\sigma}$ as the uncertainty measurement, and $\widetilde{\va}$ as a sample of scene-invariant turbulence profile.}
  \label{fig:kldseparate} 
  \vspace{-0.5em}
\end{figure*}

\noindent \textbf{The effect of the KL divergence loss}. The KL divergence regularization enforces that each entry of the sampled blur representation $\widetilde{\va}$ is i.i.d. Gaussian, which further implies that the $\widetilde{\va}$ is uncorrelated with the scene content. We observed that if we enlarge the weight of the KL divergence loss $\alpha_k$, the $\widetilde{\va}$ can be more scene-invariant. In figure \ref{fig:kldseparate}, we visualize the $\boldsymbol{\mu}$, $\boldsymbol{\sigma}$, and $\widetilde{\va}$ with a real-world static-scene sample from the BRIAR dataset \cite{cornett2023expanding} to demonstrate the meaning of the predicted LPD.

Since $\boldsymbol{\mu}$ and $\boldsymbol{\sigma}$ can be viewed as turbulence strength and uncertainty maps for the guided Mamba blocks, their quality is crucial to the performance of MambaTM. The quality is affected significantly by the weight $\alpha_k$. An experiment is conducted to demonstrate this effect, the result is shown in Table \ref{table:KLweight}. Clearly, a larger $\alpha_k$ will bring us a further performance boost. However, if the $\alpha_k$ is larger than 0.01, the training would be more unstable and the improvement starts to become marginal, so we conclude that the optimal $\alpha_k$ should be within $0.01\sim0.1$.

\begin{table}
\centering
\small
\setlength{\aboverulesep}{0pt}
\setlength{\belowrulesep}{0pt}
\resizebox{0.48\textwidth}{!}{
\begin{tabular}{lccc}
\toprule[1pt]
Benchmark & \multicolumn{2}{c}{ATSyn-static} & ATSyn-dynamic \\
 \cmidrule(r){1-1} \cmidrule(r){2-3} \cmidrule(r){4-4} 
 $\alpha_k$ & PSNR & SSIM & PSNR / SSIM / LPIPS \\
\midrule
0.001  & 27.0082  & 0.8044 & 29.9151 / 0.8843 / 0.1516 \\
0.005  & 27.1044  & 0.8036 &  29.9740 / 0.8868 / 0.1491 \\
\rowcolor{Gray} 0.01   & 27.2703  & 0.8071 &  30.0189 / 0.8883 / 0.1473 \\
\rowcolor{Gray} 0.1    & 27.2811  & 0.8069 &  30.0155 / 0.8881 / 0.1477 \\
\bottomrule[1pt]
\end{tabular}
}
\vspace{-0.5em}
\caption{Effect of the weight of KL divergence loss $\alpha_k$. }
\label{table:KLweight}
\vspace{-0.5em}
\end{table}

\subsection{Semi-supervised finetuning}
The LPD estimation facilitates the re-degradation loss during training. This implies that we can train our MambaTM model on real-world, unlabeled turbulence videos. However, simply fine-tuning the restoration model on both synthetic and real-world data will lead to the trivial solution \cite{sanghvi2024kernel}, where the restoration model just copies the input, and the turbulence estimation branch predicts zero degradation for real-world unlabeled data. Instead of the straightforward pseudo-label strategy, we employ the mean-teacher framework \cite{tarvainen2017mean} for a more stable semi-supervised fine-tuning. 

The pre-trained MambaTM is fine-tuned on the BRIAR Research dataset \cite{cornett2023expanding}, where we selected over 10000 long-range video clips that contain human subjects as the real-world data. During training, the learning rate is set to 0.0001, the batch size is 1, and the ratio between synthetic and real-world video is 1:1. The parameters of the teacher model $\theta_t$ are updated by the parameters of the student model $\theta_s$ in each step following $\theta_t = \beta \theta_t + (1-\beta)\theta_s$, where $\beta$ is set to 0.999. The model is fine-tuned by 60000 steps, then the teacher model can be viewed as the final product. To evaluate the performance of the semi-supervised fine-tuning, we restored human faces in the BRIAR Testing dataset by MambaTM before and after fine-tuning. The restored faces are then identified by a face restoration model in \cite{liu2025farsight2}. Table \ref{table:facerecognition} shows the face recognition performance with the restored face frames by different MambaTM parameters. Clearly, the simple semi-supervised fine-tuning significantly boosts the generalization capability of the restoration model.

\begin{table}
\centering
\small
\setlength{\aboverulesep}{0pt}
\setlength{\belowrulesep}{0pt}
\resizebox{0.47\textwidth}{!}{
\begin{tabular}{lccc}
\toprule[1pt]
 Rank 20 accuracy (\%) & Overall & LR Face & LR Turb \\
\midrule
Before fine-tuning & 78.4  & 85.0  & 71.4  \\
After fine-tuning  & 78.8 & 86.4 & 72.5 \\
\bottomrule[1pt]
\end{tabular}
}
\vspace{-0.5em}
\caption{Face retrieval rank 20 accuracy on the BRIAR Testing set. LR Turb, LR Face are long-range video subsets.} 
\label{table:facerecognition}
\vspace{-0.5em}
\end{table}
\section{Discussion and Conclusion}
\label{sec:conclusion}
This paper presents MambaTM, the first Mamba-based network for video turbulence mitigation, which jointly estimates and mitigates atmospheric turbulence degradation. We propose a novel latent phase distortion (LPD) representation that enhances both the efficiency and interpretability of handling turbulence. By integrating the LPD into a variational framework and employing a phase-distortion-guided Mamba block, our method efficiently enables simultaneous degradation estimation and restoration. Extensive experiments show that MambaTM delivers state-of-the-art performance with faster inference speeds, providing a robust solution for real-world video turbulence mitigation. The LPD shows a potential to learn the separation of turbulence degradation representation and scene content. However, we only considered the blur representation in this work. The estimated pixel distortion (tilt) is not scene-invariant. To facilitate better LPD for semi-supervised fine-tuning, the tilt representation should also be learned similarly.

\section{Acknowledgment}
The research is based upon work supported in part by the Intelligence Advanced Research Projects Activity (IARPA) under Contract No. 2022‐21102100004, and in part by the National Science Foundation under the grants CCSS-2030570 and IIS-2133032. The views and conclusions contained herein are those of the authors and should not be interpreted as necessarily representing the official policies, either expressed or implied, of IARPA, or the U.S. Government. The U.S. Government is authorized to reproduce and distribute reprints for governmental purposes notwithstanding any copyright annotation therein. 

{
    \small
    \bibliographystyle{ieeenat_fullname}
    \bibliography{main}
}

\clearpage
\setcounter{page}{1}
\maketitlesupplementary

\section{More details about the architecture}
\subsection{State Space Model}
To process the discrete input sequence $\boldsymbol{x}=(x_0, x_1, ..., x_{L-1})\in \mathbb{R}^{L} $, following \cite{gupta2022diagonal}, Mamba \cite{gu2023mamba} employs the zero-order hold (ZOH) assumption to convert the continuous parameters \(\boldsymbol{A}, \boldsymbol{B}\) into their discrete counterparts \(\bar{\boldsymbol{A}}, \bar{\boldsymbol{B}}\) as: $\bar{\boldsymbol{A}} = e^{\boldsymbol{\Delta} \boldsymbol{A}}$, $\bar{\boldsymbol{B}} = (\boldsymbol{\Delta} \boldsymbol{A})^{-1} (e^{\boldsymbol{\Delta} \boldsymbol{A}} - \boldsymbol{I} ) \cdot \boldsymbol{\Delta} \boldsymbol{B}$, where $\boldsymbol{\Delta}$ is the time scale. After discretizing $\boldsymbol{A}, \boldsymbol{B}$ to $\bar{\boldsymbol{A}}, \bar{\boldsymbol{B}}$, the SSM can be reformulated as:

\begin{equation}
\label{eq:SSM}
    \boldsymbol{h}_t = \boldsymbol{\bar{A}} \boldsymbol{h}_{t-1} + \bar{\boldsymbol{B}} x_t, \quad y_t = \boldsymbol{C} \boldsymbol{h}_t + \boldsymbol{D}x_t
\end{equation}

Eq.\ref{eq:SSM} represents a sequence-to-sequence mapping from \(x_t\) to \(y_t\). Since all operations are linear, all steps can be computed in parallel. To facilitate this, a convolution kernel is constructed \cite{gu_ssm}: $\mathbf{K} = (\boldsymbol{CB}, \boldsymbol{CAB}, \dots, \boldsymbol{CA}^{L-1}\boldsymbol{B})$, where the recursive multiplication of $\boldsymbol{A}$ can be efficiently computed by the scan algorithm and final output $\boldsymbol{y}$ is computed by the convolution: $\boldsymbol{y} = \boldsymbol{x} * \mathbf{K}$, which has linear complexity with respect to the length of $\boldsymbol{x}$.

However, $\mathbf{K}$ is static over time, which does not satisfy the requirement of real-world processes. To alleviate this, the selective state space model (S6) \cite{gu2023mamba} models the $\boldsymbol{\Delta}$, $\boldsymbol{B}$, $\boldsymbol{C}$ as linear projections of the input $\boldsymbol{x}$. This operation successfully enables the input-dependent selective property.

\subsection{The ReBlurNet (RBN)}
The RBN initially transforms the input image into multi-scale features, which are then modulated through element-wise multiplication with the multi-scale features of $\widetilde{\va}$ before being decoded to produce the blurred output image. While any U-Net style architecture could serve as the base network for the RBN, we ultimately selected NAFNet for this implementation. Within the RBN framework, the latent blur feature $\vb$ undergoes processing through a sequence of encoders, each comprising $1\times 1$ convolution followed by ReLU activation. The features produced by each encoder are downsampled before being passed to the subsequent encoder. We denote the output features from the four encoders as $\mathbf{e}{\vb}^{1}$, $\mathbf{e}{\vb}^{2}$, $\mathbf{e}{\vb}^{3}$, and $\mathbf{e}{\vb}^{4}$. Concurrently, the input image is processed through the base network to generate the blurred result. Importantly, before each input feature $\mathbf{v}{\vi}^{i}$ enters the $i$-th encoder for processing, it undergoes modulation via element-wise multiplication with $\mathbf{e}{\vb}^{i}$. The decoder component of the base network remains unmodified in our RBN implementation.

\section{Cost of video TM methods}
As an extension of Table 3 in the main paper, we provide the computational cost of MambaTM and other video TM methods regarding model size and MACs in Table \ref{table:cost}. Our model requires the least computation cost and has a much faster inference speed than other models. 

\begin{table}
\centering
\resizebox{0.48\textwidth}{!}{
\begin{tabular}{l|cccc}
\hline
Models & \# of params (M) & GMACs & Latency (s) \\
\hline
TSRWGAN \cite{Jin_2021_a} & 42.08 & - & 0.85 \\
TMT \cite{Zhang_2022_a} &  26.04  &  1806.0 & 0.76 \\
DATUM \cite{zhang2024spatio}  & 5.754  & 372.7 & 0.056 \\
Turb-Seg-Res \cite{saha2024turb}  & $\sim$ 30 & - & 2.404 \\
\rowcolor{Gray} MambaTM [ours] & 6.904  & 143.5 & 0.030 \\
\hline
\end{tabular}
}
\caption{The cost of different video TM methods. The GMAC and Latency are evaluated frame-wise under $960\times 540$ patches with NVIDIA A100 GPUs.}
\label{table:cost}
\end{table}

\section{Additional experiments}
\subsection{Temporal extrapolation}
Same as \cite{zhang2024spatio, Lao_2024_DTRATM}, we can also observe better performance with more input frames during testing. As shown in Table \ref{table:temporal}, our MambaTM shows good temporal extrapolation properties.

\begin{table}
\centering
\small
\setlength{\aboverulesep}{0pt}
\setlength{\belowrulesep}{0pt}
\begin{tabular}{cccc}
\toprule[0.8pt]
\# of input frames & PSNR  & SSIM &  LPIPS\\
\midrule
30  &29.5765 & 0.8793  & 0.1544 \\
40 &29.6979 & 0.8815 & 0.1530  \\
60 & 29.8129 & 0.8834 & 0.1521  \\
120 & 29.9151 & 0.8843 & 0.1516 \\ 
\bottomrule[0.8pt]
\end{tabular}
\caption{The impact of number of input frames during inference}
\label{table:temporal}
\end{table}

\begin{figure*}[t]
    \centering
    \includegraphics[width = \linewidth]{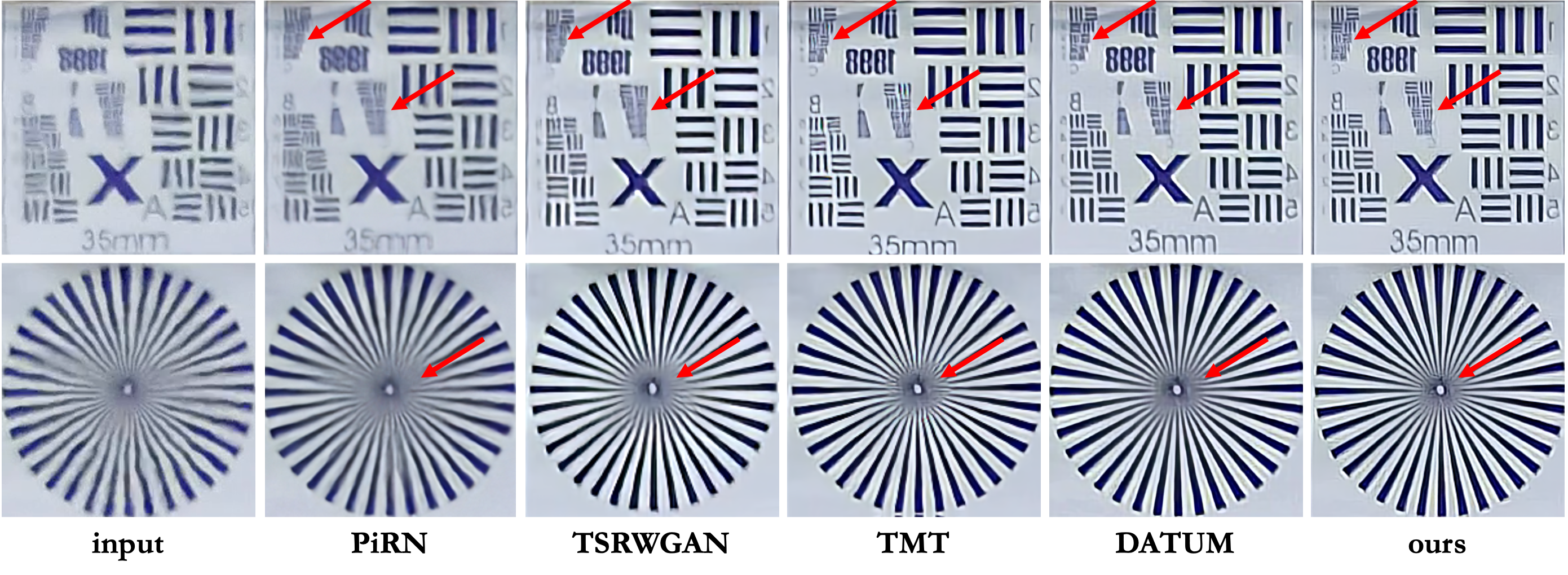}
    \caption{ualitative comparison on the OTIS dataset \cite{Gilles_2017_a}. The images on the top are from the 13th sequence, and the images on the bottom are from the 14th sequence. Zoom in for a better view.}
    \label{fig:otis}
\end{figure*}

\subsection{The latent phase distortion (LPD)}
\begin{figure}[t]
    \captionsetup[subfloat]{font=scriptsize}
    \centering
  \subfloat[clean]{%
      \includegraphics[width=0.32\linewidth]{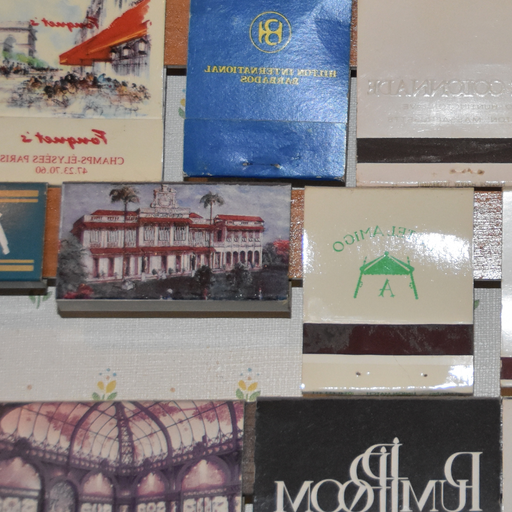}}
    \hfill
  \subfloat[degraded]{%
      \includegraphics[width=0.32\linewidth]{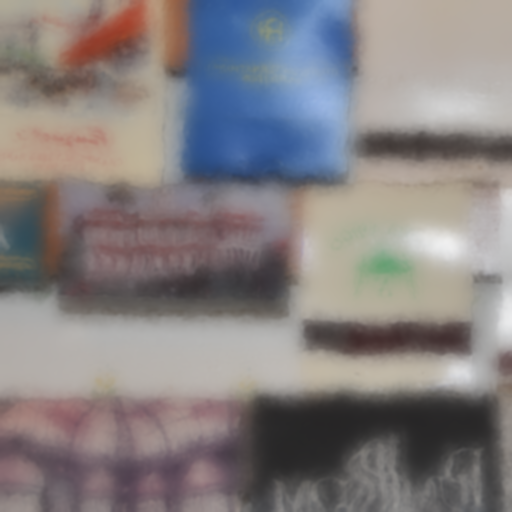}}
    \hfill
  \subfloat[re-degraded]{%
      \includegraphics[width=0.32\linewidth]{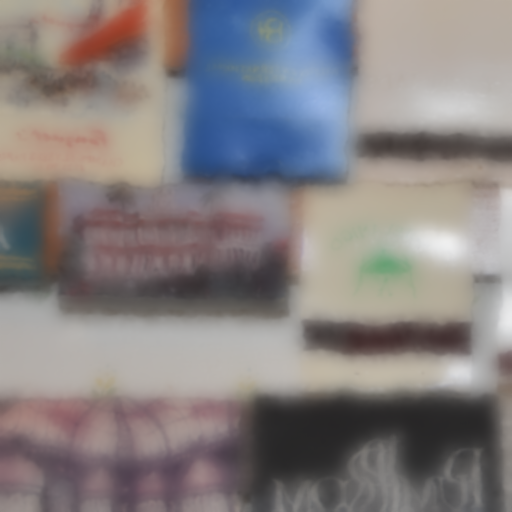}}
    \hfill
    
  \subfloat[$\mu$ ]{%
        \includegraphics[width=0.32\linewidth]{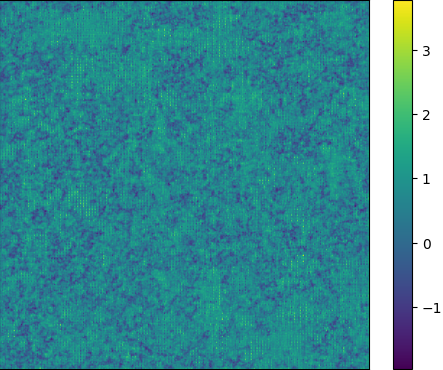}}
    \hfill
  \subfloat[$\sigma^2$ ]{%
      \includegraphics[width=0.32\linewidth]{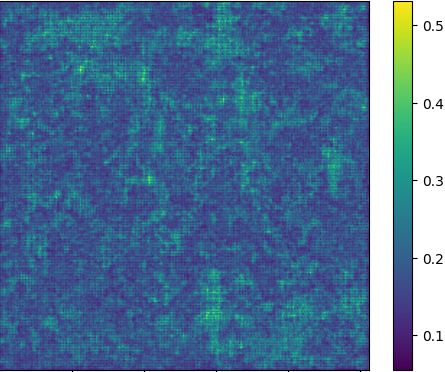}}
    \hfill
  \subfloat[$L_2$ Norm of Zernike]{%
      \includegraphics[width=0.32\linewidth]{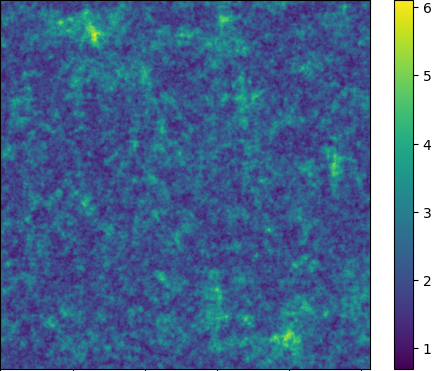}}
    \hfill
  \caption{A sample of the Zernike VAE and LPD map. (b) is generated by the Zernike-based simulator with input image (a) and Zernike coefficients whose pixel-wise norm is shown in (f), the blur kernel size is $55\times 55$. (c) is generated by our RBN with the predicted LPD, whose statistics are shown in (d) and (e). Please zoom in for a better view.}
  \label{fig:sample_vae} 
\end{figure}

We visualize an example of our Zernike VAE and LPD in Figure \ref{fig:sample_vae}. This example is taken from the validation set, featuring an unseen scene and previously unencountered turbulence parameters. We observe that the re-degraded image produced by LPD and RBN is visually similar to the degraded image generated using the Zernike coefficients. The mean of LPD, $\mu$, represents the overall turbulence strength, while the variance $\sigma^2$ is visually correlated with the local variation, as indicated by the pixel-wise $L_2$ norm of the corresponding Zernike coefficients.

\subsection{Real-world samples of the LPD-based simulation}
\begin{figure}[t]
    \centering
    \includegraphics[width = \linewidth]{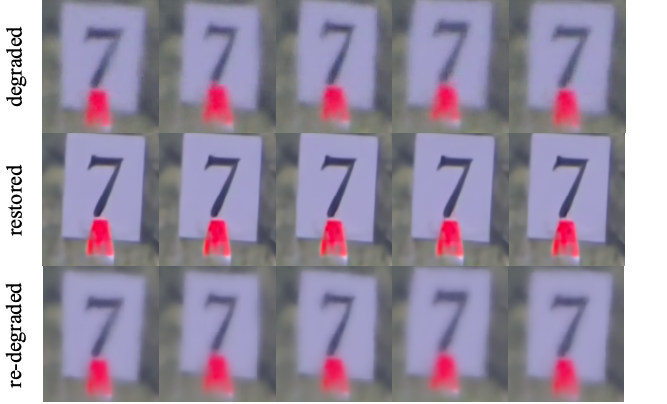}
    \caption{Comparing the real-world turbulent images (from \emph{BRIAR\_1} in the supplementary material) and re-degraded images.}
    \label{fig:realworldreturb}
\end{figure}

To demonstrate the generalization capability of the LPD estimation and our LPD-based simulator, we provide a real-world testing case in Figure \ref{fig:realworldreturb}. It can be seen that our model successfully recovered the clean patterns from the turbulence-affected images across a long-range distance. By comparing the real-world degraded and our re-degraded images using the restored image as the input, we can find that our simulator can faithfully represent real-world turbulence. We also provide the associated videos in the supplementary material. 

\subsection{More qualitative comparison}
\begin{figure*}[ht]
    \captionsetup[subfloat]{font=scriptsize}
    \centering
  \subfloat[Input frame]{%
      \includegraphics[width=0.49\linewidth]{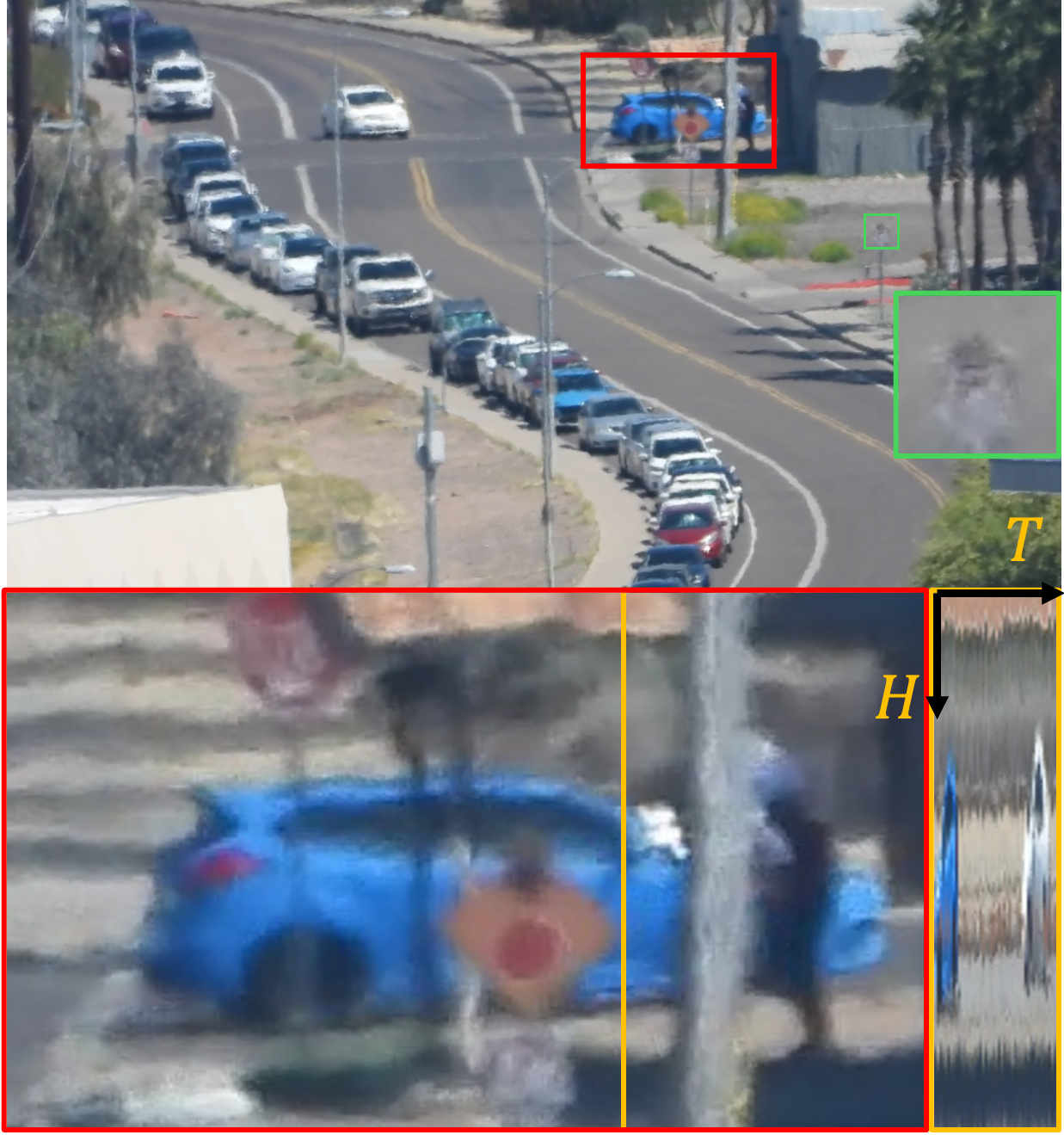}}
    \hfill
  \subfloat[Turb-Seg-Res (CVPR 2024) \cite{saha2024turb}]{%
      \includegraphics[width=0.49\linewidth]{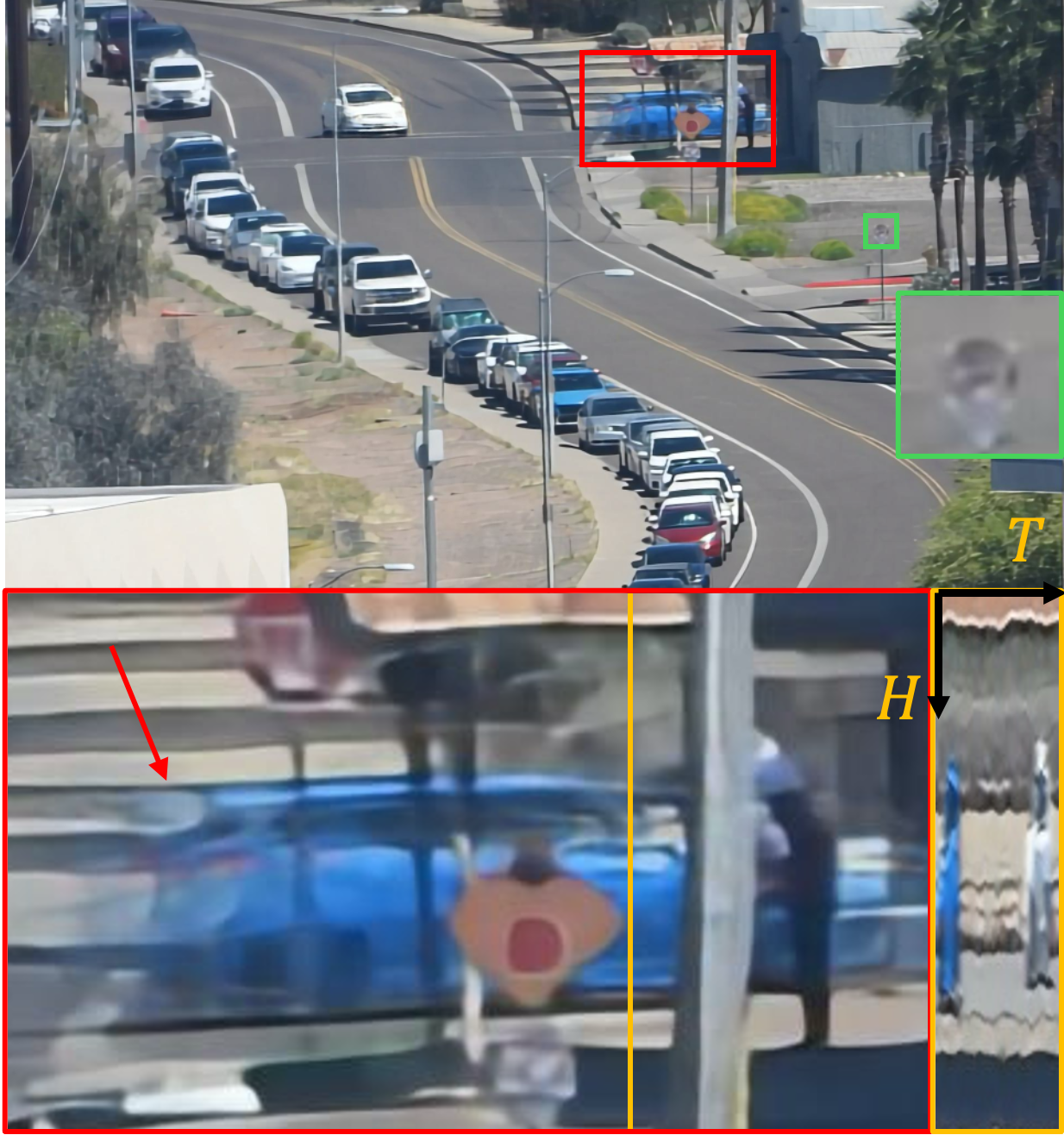}}
    \hfill

  \subfloat[DATUM (CVPR 2024) \cite{zhang2024spatio}]{%
        \includegraphics[width=0.49\linewidth]{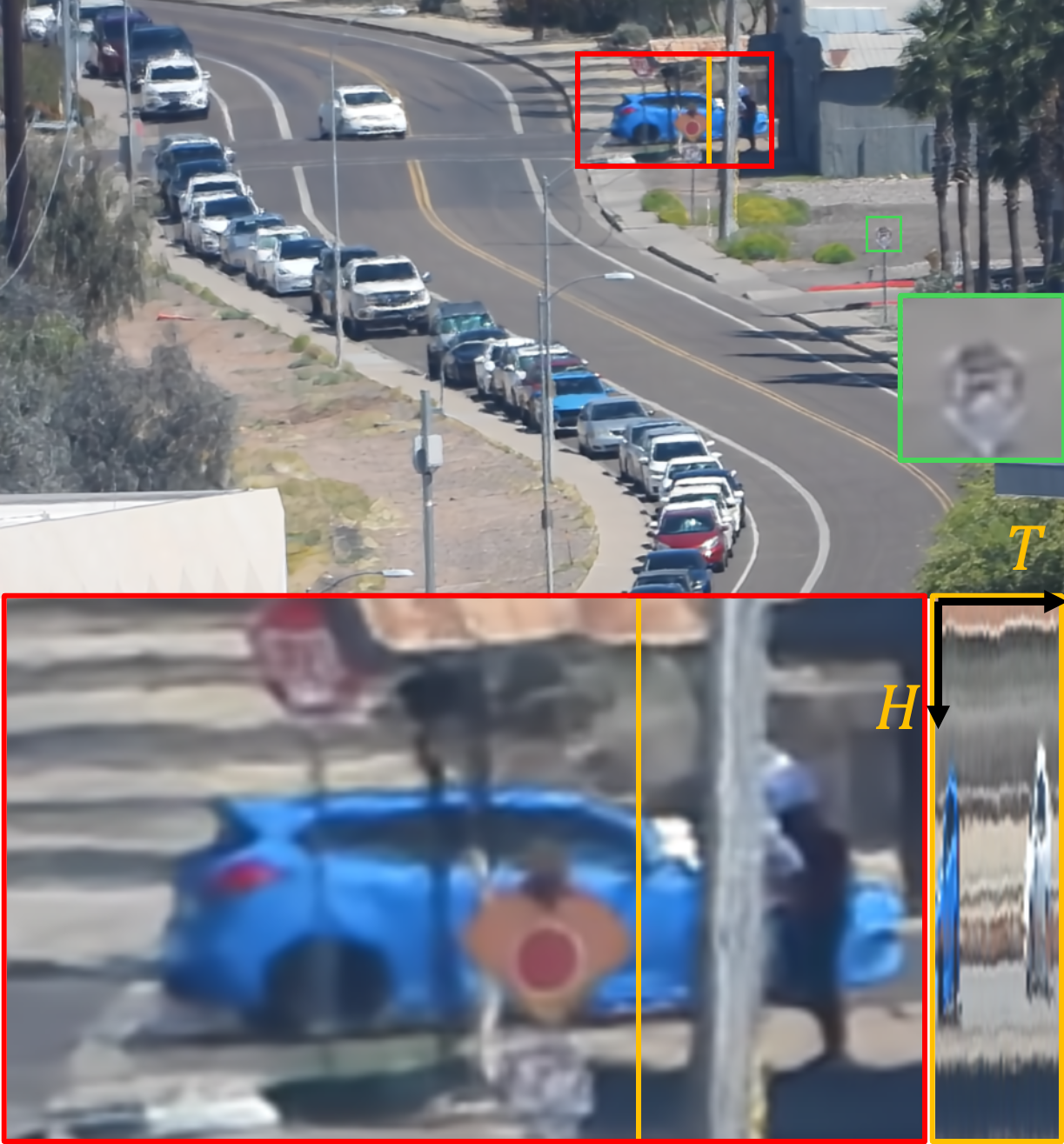}}
    \hfill
  \subfloat[MambaTM (ours)]{%
      \includegraphics[width=0.49\linewidth]{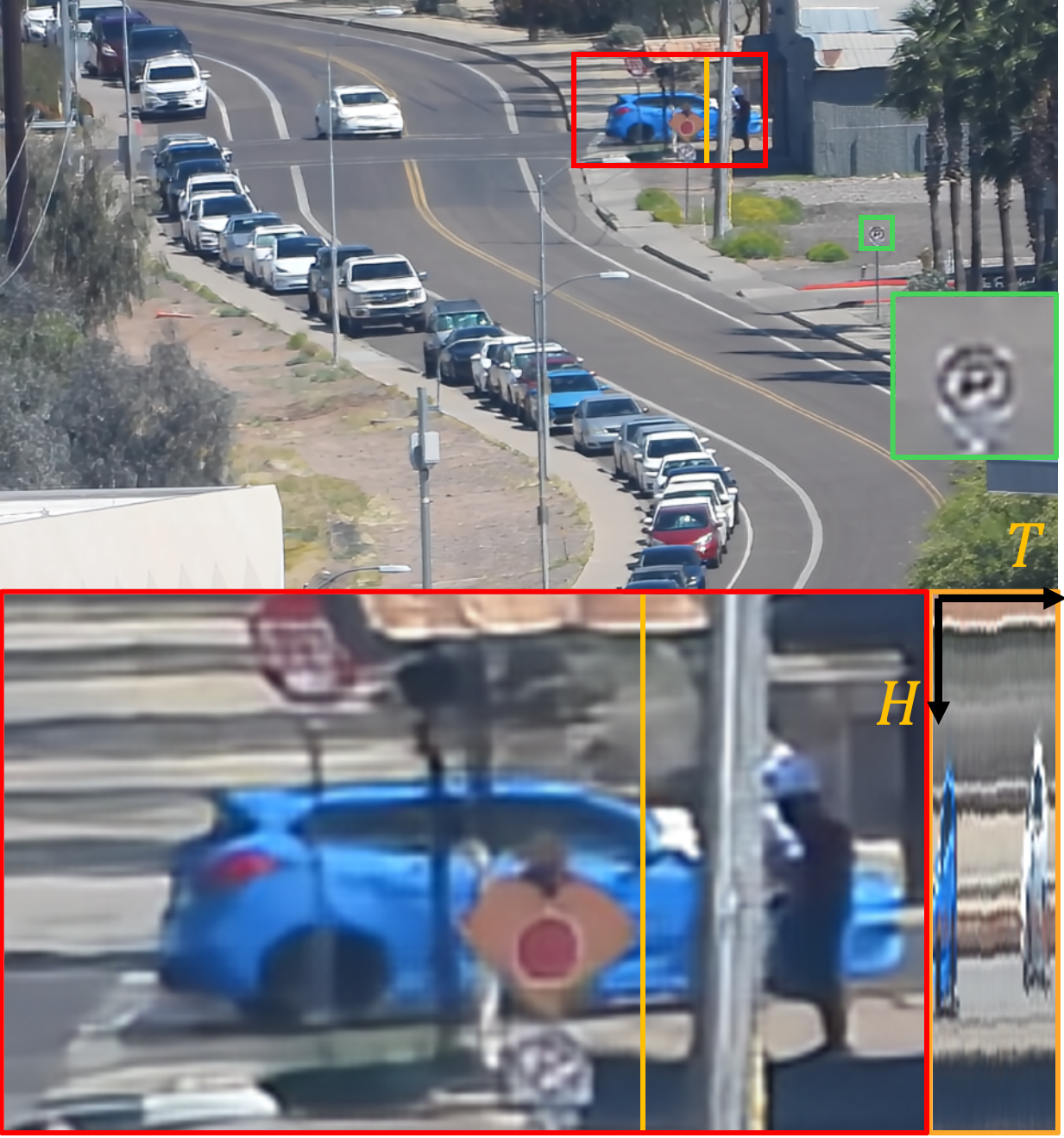}}
    \hfill
  \caption{Qualitative comparison on the URG-T real-world dataset \cite{saha2024turb}. From the green box, we can find that \emph{spatially}, our method can produce the sharpest and most reliable restoration. We provide temporal slices (the orange line in red bounding boxes of each frame) in the bottom right of each figure, from which we can find that \emph{temporally}, our method generates the most stable and consistent output. Note that Figure (b) also suffers from the ghost effect caused by its temporal fusion method.}
  \label{fig:car} 
\end{figure*}

To demonstrate the advancement of our method, we further provide two real-world comparisons. The first is on the static scenes from the OTIS dataset \cite{Gilles_2017_a}. As presented in Figure \ref{fig:otis}, we compare MambaTM with other SOTA turbulence mitigation works and we can find that our method recovers more details than others. The second is on the dynamic scene from the URG-T dataset \cite{saha2024turb}, we compare MambaTM with two recent SOTA DATUM \cite{zhang2024spatio} and Turb-Seg-Res \cite{saha2024turb}. To highlight our method's temporal consistency on dynamic scenes, we fetch 1D spatial slices from the same location in each frame of the image sequences and stitch all slices along the time axis. The result is shown in Figure \ref{fig:car}. From this, we can find that our method shows better restoration quality both spatially and temporally. Meanwhile, notice that our method is $2\times$ faster than DATUM and $50\times$ faster than Turb-Seg-Res.

\end{document}